\documentclass[11pt]{article}

\usepackage[final]{acl}

\usepackage{times}
\usepackage{latexsym}

\usepackage[T1]{fontenc}
\usepackage[utf8]{inputenc}

\usepackage{microtype}

\usepackage{inconsolata}

\usepackage{graphicx}
\usepackage{booktabs}

\title{\textit{BhashaSutra}: A Task-Centric Unified Survey of Indian NLP Datasets, Corpora, and Resources}

\author{
Raghvendra Kumar$^{1}$ \quad
Devankar Raj$^{2}$ \quad
Sriparna Saha$^{1}$ \\
\\
$^{1}$Department of Computer Science and Engineering, Indian Institute of Technology Patna, India \\
$^{2}$Indian Institute of Technology Patna, India \\
\\
\texttt{devankarraj@gmail.com}\quad\quad\texttt{\{raghvendra\_2221cs27,sriparna\}@iitp.ac.in}
}

\begin{document}
\maketitle

\begin{abstract}
India’s linguistic landscape, spanning 22 scheduled languages and hundreds of marginalized dialects, has driven rapid growth in NLP datasets, benchmarks, and pretrained models. However, no dedicated survey consolidates resources developed specifically for Indian languages. Existing reviews either focus on a few high-resource languages or subsume Indian languages within broader multilingual settings, limiting coverage of low-resource and culturally diverse varieties. To address this gap, we present the first unified survey of Indian NLP resources, covering \textbf{200+ datasets}, \textbf{50+ benchmarks}, and \textbf{100+ models, tools, and systems} across text, speech, multimodal, and culturally grounded tasks. We organize resources by linguistic phenomena, domains, and modalities; analyze trends in annotation, evaluation, and model design; and identify persistent challenges such as data sparsity, uneven language coverage, script diversity, and limited cultural and domain generalization. This survey offers a consolidated foundation for equitable, culturally grounded, and scalable NLP research in the Indian linguistic ecosystem.
\end{abstract}

\section{Introduction}

India hosts one of the world’s most linguistically diverse ecosystems, with \textbf{22 scheduled languages} and hundreds of dialects spanning multiple scripts and language families. Several Indian languages such as Hindi, Bengali, Telugu, Marathi, Tamil, Urdu and Gujarati, rank among the \textbf{most spoken languages globally}, collectively serving hundreds of millions of speakers\footnote{\url{https://en.wikipedia.org/wiki/List_of_languages_by_number_of_native_speakers}}. This scale and diversity make Indian languages both scientifically important and socially consequential for NLP research.

In recent years, Indian-language NLP has witnessed rapid growth, with datasets, benchmarks, and pretrained models emerging across domains including healthcare, law, education, governance, finance, and media. However, progress remains fragmented: most efforts focus on a few relatively high-resource languages, exhibit wide variation in quality and documentation, and are scattered across venues. Existing surveys either target narrow task families or embed Indian languages within broad multilingual settings, leaving no unified, task-comprehensive overview dedicated exclusively to Indian NLP \cite{kakwani2020indicnlpsuite,panchal2024nlp,lahoti2022survey,kalamkar2021benchmarks,harish2020comprehensive,kumar2022indicnlg,khan2024indicllmsuite}.

\textbf{This survey addresses this gap.} We organize the Indian NLP landscape into \textbf{six high-level groups comprising seventeen fine-grained tasks}: 
(i) \emph{\textbf{Core Linguistic Processing}}: tokenization, normalization, and morphological analysis; POS tagging; named entity recognition; 
(ii) \emph{\textbf{Text Classification and Semantics}}: sentiment and emotion analysis; hate speech and toxicity detection; topic classification; natural language understanding; 
(iii) \emph{\textbf{Generation and Translation}}: summarization; machine translation; question answering; 
(iv) \emph{\textbf{Retrieval and Interaction}}: information retrieval; dialogue systems; 
(v) \emph{\textbf{Speech and Multimodality}}: speech processing; multimodal language understanding; and 
(vi) \emph{\textbf{Societal, Cultural, and Emerging Tasks}}: misinformation and fact-checking; cultural reasoning; and other emerging tasks. \textbf{The key contributions of this survey are as follows:}

\emph{
$\triangleright$ A \textbf{unified, task-centric taxonomy} of Indian NLP covering seventeen tasks across text, speech, and multimodal settings.\\
$\triangleright$ A \textbf{comprehensive consolidation of datasets, benchmarks, and tools/systems}, highlighting language coverage, resource imbalance, and evaluation practices.\\
$\triangleright$ A focused analysis of \textbf{societal and cultural challenges}, including misinformation, cultural reasoning, bias, and code-mixing, central to the Indian context.\\
$\triangleright$ Identification of \textbf{open gaps and future research directions} toward scalable, inclusive, and trustworthy Indian-language NLP.
}

\begin{figure*}[!htbp]
    \centering
    \includegraphics[width=0.9\textwidth]{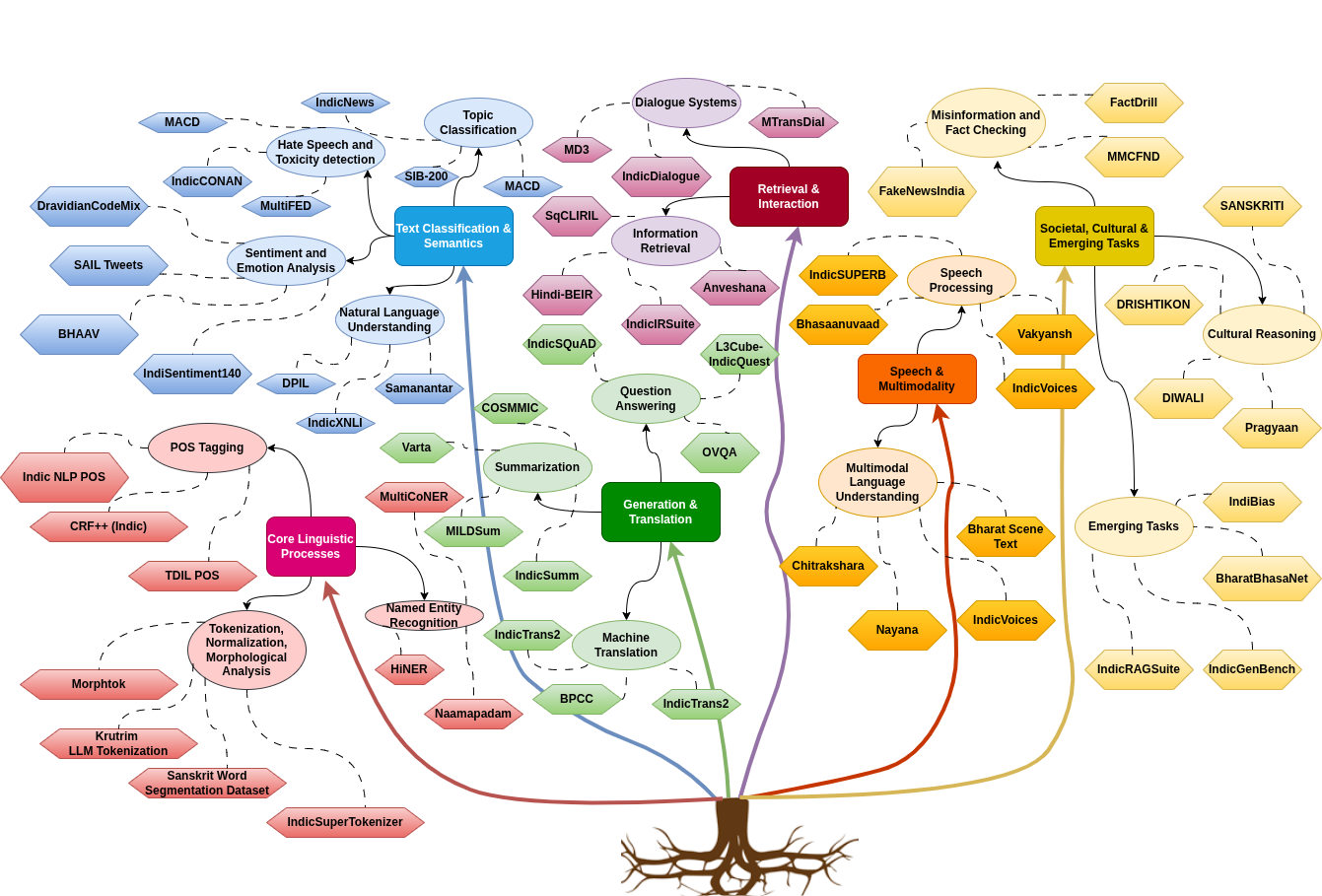}
    \caption{Task-centric organization of Indian NLP resources. The figure presents six high-level task branches—\emph{Core Linguistic Processes}, \emph{Text Classification \& Semantics}, 
    \emph{Generation \& Translation}, \emph{Retrieval \& Interaction}, \emph{Speech \& Multimodality}, and 
    \emph{Societal, Cultural \& Emerging Tasks}. Each branch is further decomposed into constituent subtasks, which are illustrated using representative datasets, benchmarks, and tools selected to reflect methodological and resource diversity rather than completeness or prominence. The diagram highlights the structural relationships between tasks and resources across the Indian NLP ecosystem.
    }\label{fig:task_tree}
\end{figure*}

\begin{figure*}[htbp]
    \centering
    \includegraphics[width=0.9\linewidth]{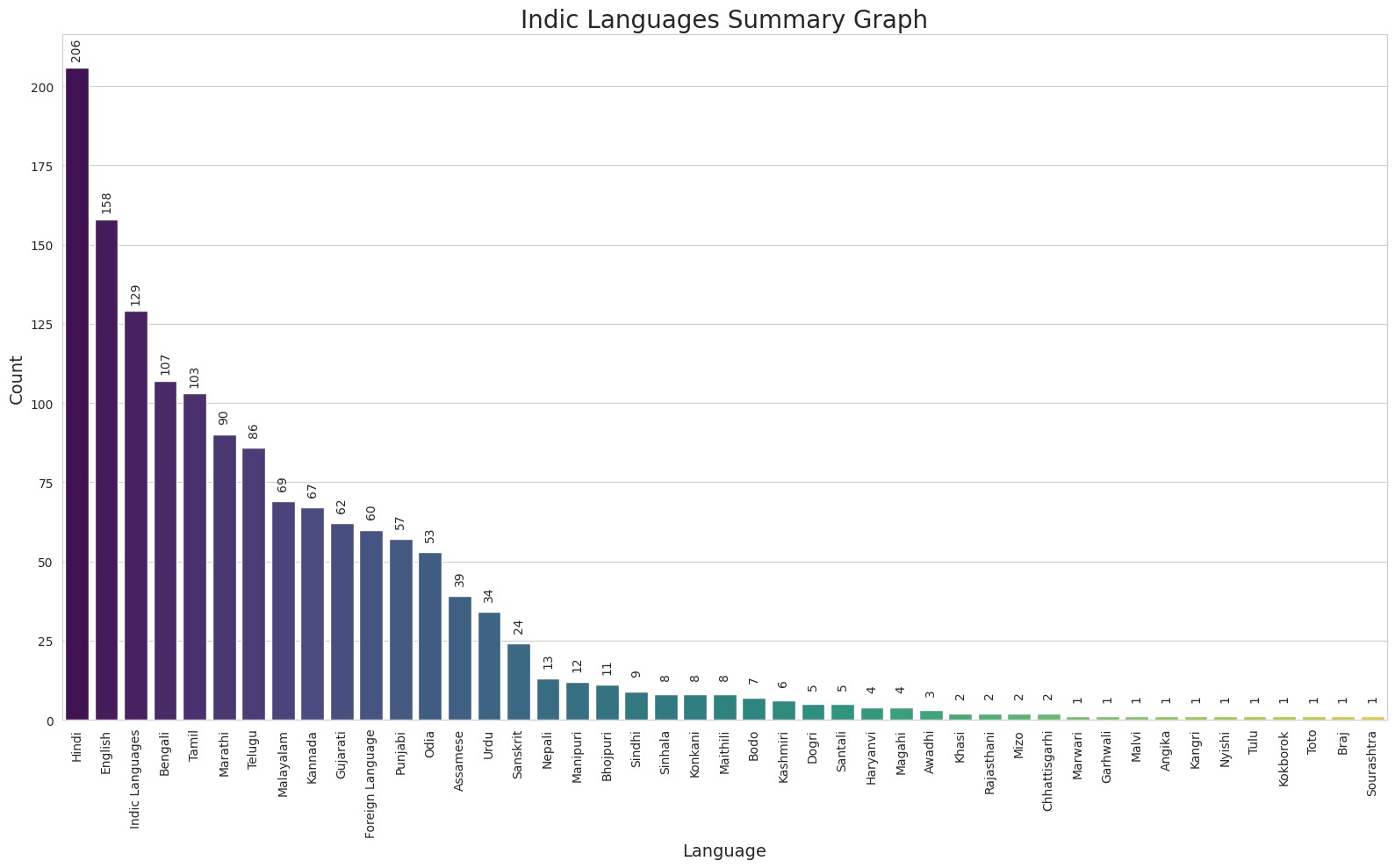}
    \caption{
    Language-wise distribution of datasets and studies across Indian NLP tasks. Single-language resources are counted individually, while multilingual resources are grouped under \emph{Indic Languages}. This aggregation aids visualization but may mask uneven coverage, often favoring higher-resource languages. We emphasize that these counts reflect the breadth of research activity and task coverage, rather than aggregate dataset size or volume.
    }
    \label{fig:master}
\end{figure*}

For each task, we summarize key datasets, benchmarks, and tools for Indian languages, including multilingual resources with English. Each subsection provides a concise snapshot of representative approaches, while cross-cutting gaps are discussed in Appendix~\ref{sec:unified_gaps}. The appendix also includes task-wise resource tables and language-wise distributions. Figure~\ref{fig:task_tree} presents a task-centric overview, and Figure~\ref{fig:master} summarizes language-wise resource counts.

To clarify scope, inclusion criteria, and categorization choices, we provide a concise FAQ in Appendix~\ref{sec:faq}. Resources were identified through systematic searches across major NLP venues (e.g., ACL, EMNLP, NAACL, COLING, LREC, Interspeech), arXiv, and institutional repositories (e.g., AI4Bharat, LDC-IL, IIITH-ILSC), complemented by citation chaining and task-specific keyword queries; detailed screening procedures are described in Appendix~\ref{sec:resource_collection}.

We do not assign explicit quality rankings, as evaluation standards vary across tasks and modalities. Instead, we report dataset characteristics and documented limitations to enable assessment of resource suitability.

In addition to consolidation, we provide cross-task synthesis of recurring ecosystem-level challenges, including language imbalance, annotation fragmentation, domain skew, evaluation inconsistency, and cross-lingual brittleness. While the resource landscape continues to evolve, the proposed taxonomy and gap analysis are intended as a stable and extensible framework for future Indian NLP research. Where available, we also report dataset licensing and usage constraints, mentioned in Appendix~\ref{sec:licensing}.

\section{Core Linguistic Processing}

\subsection{Tokenization, normalization, and morphological analysis}
\label{subsec:tokenization}

Tokenization, normalization, and morphological analysis are foundational for Indian-language NLP, where rich morphology, diverse scripts, and sandhi limit the effectiveness of generic subword methods. \textbf{Tokenization research} includes morphology-aware approaches such as Morphtok \cite{brahma2025morphtok}, studies on low-resource languages like Santhali \cite{ohm2024study}, and evidence of downstream gains in tasks such as zero-shot NER \cite{pattnayak2025tokenization}. Large-scale multilingual efforts, including Krutrim LLM \cite{kumar2024krutrim} and IndicSuperTokenizer \cite{rana2025indicsupertokenizer}, propose Indic-centric tokenizer designs, while toolkits such as iNLTK \cite{arora2020inltk} support practical normalization. 

\textbf{Normalization and lexical processing} are further addressed through improvements to Bengali and Hindi Large Language Models (LLMs) \cite{shahriar2024improving}, word embeddings \cite{saurav2020passage}, word similarity resources \cite{akhtar2017word}, and punctuation and inverse text normalization via indic-punct \cite{gupta2022indic}. \textbf{Morphological analysis} spans resources for Sanskrit segmentation and parsing \cite{krishnan2025normalized,krishna2017dataset}, Gujarati analyzers \cite{baxi2022gujmorph,baxi2025bidirectional}, Malayalam and Tamil systems \cite{premjith2018deep,rajasekar2021comparison,sarveswaran2018thamizhifst}, Telugu analyzers \cite{dasari2023transformer}, Punjabi morphological evaluation \cite{singh2021morphological}, multiword expression datasets \cite{singh2016multiword}, and early statistical analyzers \cite{srirampur2014statistical,prathibha2013development}. Additional details are provided in the Appendix (Table~\ref{tab:tokenization_morphology} and Figure~\ref{fig:language_count_tokenization}).

\subsection{Part-of-Speech (POS) Tagging}
\label{subsec:pos}

POS tagging for Indian languages has been studied using classical models such as trigram HMMs \cite{sarkar2013trigram} and CRF-based systems for Odia \cite{dalai2023part}, as well as neural approaches including deep models for South Indian languages \cite{rajani2022pos}, character-level architectures for Assamese \cite{phukan2024exploring}, transformer-based taggers for Odia \cite{dalai2024deep}, and unsupervised deep tagging for Sanskrit \cite{srivastava2018deep}. Data-scarce settings are addressed through methods for extremely low-resource languages \cite{kumar2024part}, cross-lingual tagging using related-language resources \cite{reddy2011cross}, and comparative studies for Magahi \cite{kumar2012developing}, supported by corpora and resources such as Bengali news and lexicon-derived datasets \cite{ekbal2008web,dash2013part} and unified parsing proposals \cite{tandon2017unity}. Additional details are provided in the Appendix (Table~\ref{tab:pos_indian_languages} and Figure~\ref{fig:pos}).

\subsection{Named Entity Recognition}
\label{subsec:ner}

Named Entity Recognition (NER) has been extensively studied for Indian languages, encompassing both dataset creation and model development. Resource efforts include large-scale datasets such as HiNER for Hindi \cite{murthy2022hiner}, Naamapadam for multiple Indic languages \cite{mhaske2023naamapadam}, Marathi \cite{litake2022l3cube}, Bangla \cite{haque2023b}, Assamese \cite{pathak2022asner}, and early Sinhala corpora \cite{dahanayaka2014named}, alongside low-resource datasets for Bhojpuri, Maithili, and Magahi \cite{mundotiya2023development} and pre-annotated Sanskrit resources \cite{sujoy2023pre}. Modeling approaches span CRF and neural systems for Hindi \cite{sharma2022named,sharma2020deep}, bilingual and embedding-enhanced models for Hindi–Punjabi \cite{goyal2021deep}, multilingual transformer fine-tuning \cite{bahad2024fine,mohan2023fine}, efficient architectures for Punjabi \cite{singh2023deepspacy}, and broader multilingual benchmarks such as MultiCoNER \cite{malmasi2022multiconer}. Further details are provided in the Appendix (Table~\ref{tab:indic_ner_resources} and Figure~\ref{fig:ner}).

\section{Text Classification and Semantics}

\subsection{Sentiment and Emotion Analysis}
\label{subsec:sentiment}

Sentiment and emotion analysis for Indian languages spans dataset creation, code-mixing, multimodality, and cross-lingual transfer. Datasets include multilingual and low-resource resources such as IndiSentiment140 \cite{kumar2024indisentiment140}, SAIL shared-task tweets \cite{patra2015shared,phani2016sentiment}, Malayalam–English and Kannada–English code-mixed corpora \cite{chakravarthi2020sentiment,hande2020kancmd,kannadaguli2021code}, Marathi and Hindi datasets \cite{kulkarni2021l3cubemahasent,ekbal2022hindimd}, Bangla–English–Hindi test sets \cite{raihan2023sentmix}, and classical resources covering Hindi, Telugu, Odia, and Bhagavad Gita translations \cite{akhtar2016aspect,regatte2020dataset,naidu2017sentiment,sahu2016sentiment,chandra2022semantic}. Additional code-mixed and Dravidian resources include DravidianCodeMix \cite{chakravarthi2022dravidiancodemix}, EmoInHindi \cite{singh2022emoinhindi}, and Hinglish emotion datasets \cite{sasidhar2020emotion,patra2018sentiment}. Multimodal corpora span DravidianMultimodality \cite{chakravarthi2021dravidianmultimodality} and Marathi emotion datasets \cite{chaudhari2023mahaemosen}. Emotion-focused corpora include Anubhuti \cite{pal2020anubhuti}, Bhaav \cite{kumar2019bhaav}, Navrasa \cite{saini2020kavi}, and lyrical-text datasets \cite{dhar2025emotion}, with speech-based resources such as IITKGP-SESC/SEHSC \cite{koolagudi2009iitkgp,koolagudi2011iitkgp} and South Indian emotion corpora \cite{poorna2018weight}. Modeling approaches span CNNs \cite{gupta2021toward,shalini2018sentiment}, embeddings and transfer learning \cite{ahmad2020borrow}, transformer-based and cross-lingual methods \cite{kumar2021sentiment,kumar2023zero}, and multimodal transformers \cite{kumar2025indidataminer}. Additional details are provided in the Appendix (Table~\ref{tab:sentiment_emotion_indic_appendix} and Figure~\ref{fig:senti}).

\subsection{Hate Speech and Toxicity Detection}
\label{subsec:hate}

Hate speech and toxicity detection for Indian languages spans multilingual, code-mixed, domain-specific, and multimodal settings, with emphasis on low-resource and socio-cultural contexts. Datasets include target-based Hindi hate speech (TABHATE) \cite{sharma2024tabhate}, Hindi–English code-mixed corpora \cite{bohra2018dataset,gupta2021hindi,sreelakshmi2020detection}, multilingual Indic resources such as IndicConan \cite{sahoo2024indicconan}, caste-based hate datasets \cite{gupta2025caste}, Assamese \cite{ghosh2023transformer}, Bengali \cite{romim2021hate,mondal2024detecting}, Marathi (L3Cube-MahaHate) \cite{velankar2022l3cube}, Odia and Dravidian datasets \cite{roy2022hate,sreelakshmi2024detection,anbukkarasi2023deep}, Telugu corpora \cite{khanduja2024telugu}, Indo-Aryan resources \cite{narayan2023hate}, election-domain datasets such as CHUNAV \cite{jafri2024chunav}, and large-scale multilingual abusive-comment corpora \cite{gupta2022multilingual,jhaveri2022toxicity}. Modeling approaches include translation-based detection \cite{biradar2021hate}, transformer and deep-learning systems \cite{ghosh2023transformer,velankar2021hate,kapil2023hhsd}, ensemble and multitask learning \cite{sharma2025stop,ghosal2023hatecircle}, federated learning \cite{singh2024generalizable}, bootstrapping for low-resource settings \cite{das2022data,gupta2022adima}, lightweight LLM adaptation \cite{aloiso2024lightweight}, multimodal Dravidian hate detection \cite{anilkumar2024dravlangguard}, video-based toxicity modeling \cite{maity2024toxvidlm}, and unified multilingual solutions \cite{bhatia2021one,beniwal2025unityai}.

\subsection{Topic Classification and Document Categorization}
\label{subsec:topic}

Topic classification assigns thematic or domain labels to text and supports large-scale content organization in Indian languages. Multilingual and regional news datasets include L3Cube-IndicNews \cite{mirashi2023l3cube} and low-resource Kashmiri benchmarks \cite{deyar2025dataset}. Domain- and snippet-level resources span Telugu headline classification \cite{kanumolu2024teclass}, factual-claim detection in Indian social media \cite{dutta2022multilingual}, Tamil meme and troll classification \cite{suryawanshi2020dataset}, and Telugu social-media categorization \cite{dalal2023mmt}. Language-agnostic and tool-supported approaches leverage monolingual corpora such as AI4Bharat IndicNLP \cite{kunchukuttan2020ai4bharat} and efficient multilingual classification methods \cite{aggarwal2021efficient,ramraj2020topic}. A detailed comparison is provided in Appendix (Table~\ref{tab:topic_classification_indic} and Figure~\ref{fig:topic}).

\subsection{Natural Language Understanding}
\label{subsec:nlu}

Natural Language Understanding (NLU) for Indian languages spans paraphrasing, inference, and semantic similarity, with a focus on resource-intensive tasks. Paraphrasing includes early neural, rule-based, and unsupervised methods \cite{bhargava2017deep,sethi2016novel,praveena2017chunking,singh2021augmented,bhole2018detection,das2018unsupervised}, multilingual corpora \cite{singh2020creating}, and language-specific datasets and models for Marathi, Kannada, Bangla, Telugu, Sanskrit, and Punjabi \cite{jadhav2025mahaparaphrase,anagha2023paraphrase,gupta2025knparaphraser,akil2022banglaparaphrase,rohith2022telugu,saha2024bnpc,dhingra2022rule,singh2020deep}. Inference research includes multilingual and code-mixed Natural Language Inference (NLI) datasets \cite{aggarwal2022indicxnli,khanuja2020new}, large evaluation suites \cite{ahuja2023mega,kudugunta2023madlad}, and Indic-focused resources such as BanglaBERT, IndicIRS-Suite, bilingual tabular inference, Hindi RC, and MILU \cite{bhattacharjee2022banglabert,haq2024indicirsuite,agarwal2022bilingual,anuranjana2019hindirc,verma2025milu}. Semantic similarity work spans word- and sentence-level datasets \cite{akhtar2017word,pandit2019improving,mirashi2025l3cube,chandrashekar2024fasttext}, cognate and false-friend evaluations \cite{kanojia2020challenge}, and embedding-based studies \cite{yadav2024semantic}, supported by parallel and comparable corpora for cross-lingual transfer \cite{ramesh2022samanantar,soni2021dataset,saurav2020passage,siripragada2020multilingual}. Additional details are provided in Appendix (Table~\ref{tab:nlu-indic-resources} and Figure~\ref{fig:nlu}).

\section{Generation and Translation}

\subsection{Summarization}
\label{subsec:summarization}

Summarization for Indian languages covers abstractive, extractive, multilingual, multimodal, and conversational settings across legal, news, social-media, and low-resource domains. Representative datasets include MILDSum \cite{datta2023mildsum}, IndicSumm \cite{sireesha2023indicsumm}, PMIndiaSum \cite{urlana2023pmindiasum}, MahaSum \cite{kulkarni2024l3cube}, HindiSumm \cite{singh2024hindisumm}, Social-Sum-Mal \cite{rahul2024social}, TeSum \cite{urlana2022tesum}, M3LS \cite{verma2023large}, Gupshup \cite{mehnaz2021gupshup}, COSMMIC \cite{kumar2025cosmmic}, multimodal discussion summarization \cite{singh2025conversations}, and low-resource or specialized datasets for Konkani \cite{d2019development,d2022automatic}, Gujarati \cite{mehta2022automatic}, Urdu \cite{raza2024end}, Tamil speech \cite{nithyakalyani2019speech}, and regional headline generation \cite{madasu2023mukhyansh}. Large-scale benchmarks such as Varta \cite{aralikatte2023varta} and IndicGenBench \cite{singh2024indicgenbench} broaden evaluation coverage. Modeling spans transformer-based abstractive systems \cite{kulkarni2024l3cube,ghosh2024medsumm}, embedding-driven and evolutionary methods \cite{khan2025automatic,jain2022automatic}, neural Punjabi and Malayalam models \cite{jain2021text,k2023abstractive}, low-resource LM-based approaches \cite{munaf2024low,kumar2026comments}, multimodal and multilingual fine-tuning \cite{phani2024mmsft,mane2024amalgam,kumar2024indicbart,ghosh2024healthalignsumm}, LLM-based regional summarization \cite{sawant2024saralmarathi}, and dialogue summarization with mT5 and IndicBART \cite{sharma2024evaluating}. Full resource details are provided in Appendix (Table~\ref{tab:indic_summarization_resources}, Figure~\ref{fig:summ}).

\subsection{Machine Translation}
\label{subsec:mt}

Machine Translation (MT) for Indian languages spans bilingual, multilingual, and domain-specific settings. Core parallel corpora include Samanantar and related collections \cite{siripragada2020multilingual,haddow2020pmindia}, Indo-Aryan and Dravidian datasets \cite{baruah2021low,choudhary2020neural}, recent expansions via CorIL \cite{bhattacharjee2025coril}, and large-scale multilingual bitext supporting all 22 scheduled languages through IndicTrans2 \cite{gala2023indictrans2}. Domain-specific and language-pair resources cover governance \cite{mujadia2025ilgov}, legal \cite{mahapatra2025milpac}, education \cite{appicharla2021edumt}, speech translation \cite{jain2024bhasaanuvaad}, Sanskrit--English/Hindi \cite{sethi2023novel,maheshwari2024samayik}, Marathi--English \cite{jadhav2020marathi}, and multilingual augmentation via EnIndic \cite{banerjee2023automatic}. 

Recent systems predominantly adopt transformer-based NMT, improving seq2seq baselines for low-resource and Indic--Indic pairs, including Assamese--Indic \cite{baruah2021low}, Sanskrit--Hindi \cite{sethi2023novel}, Tamil--English \cite{jain2020neural,choudhary2018neural}, Kannada--English \cite{nagaraj2021kannada}, and extremely low-resource settings \cite{lalrempuii2023extremely,bisht2024neural,bala2024multilingual,suman2023iacs}. Applied MT spans e-commerce \cite{patil2022large}, poetry translation \cite{chakrawarti2022machine}, and multimodal MT \cite{parida2019hindi}, supported by transliteration and evaluation resources such as Aksharantar \cite{madhani2023aksharantar}, MuRIL \cite{khanuja2021muril}, and IndicMT Eval \cite{dixit2023indicmt}. Full details are provided in Appendix (Table~\ref{tab:indic_mt}, Figure~\ref{fig:mt}).

\subsection{Question Answering}
\label{subsec:qa}

Question Answering (QA) for Indian languages spans extractive, generative, structured, multimodal, and long-form settings. Monolingual and language-specific resources include TransQAM for Malayalam \cite{rahmath2025transqam}, KrishiQ-BERT for Kannada \cite{ajawan2024krishiq}, Marathi QA \cite{amin2023question}, Hindi–Marathi QA \cite{sabane2023breaking}, TeQuAD for Telugu \cite{vemula2022tequad}, MahaSQuAD for Marathi \cite{ghatage2024mahasquad,ghatage2023mahasquad}, Bengali factoid QA \cite{das2022improvement}, Sanskrit kāraka-based QA \cite{verma2023karaka}, Sinhala QA \cite{ranasinghe2025question}, culturally grounded CaLMQA \cite{arora2025calmqa}, and open-domain Telugu QA \cite{ravva2020avadhan}. Multilingual and cross-lingual efforts include MLQA \cite{lewis2020mlqa}, BharatBBQ \cite{tomar2025bharatbbq}, MMQA \cite{gupta2018mmqa}, MUCOT \cite{kumar2022mucot}, structured QA with state-space models \cite{vats2025multilingual}, EHMMQA \cite{lahoti2025ehmmqa}, and long-context QA \cite{mishra2025long}.

Multimodal QA resources include OVQA for Odia \cite{parida2025ovqa}, Indic VQA \cite{chandrasekar2022indic}, handwritten multilingual VQA \cite{pal2025hw}, Assamese AVQA \cite{rahman2024tdiuc}, and Tamil grammar QA via knowledge graphs \cite{mithilesh2024aganittyam}. Unified evaluation is supported by IndicSQuAD \cite{endait2025indicsquad}, Indic table QA \cite{pal2024table}, the Indic QA Benchmark \cite{singh2025indic}, and L3Cube-IndicQuest \cite{rohera2024l3cube}. A comprehensive overview is provided in Appendix (Table~\ref{tab:indic_qa_overview} and Figure~\ref{fig:qa}).

\section{Retrieval and Interaction}

\subsection{Information Retrieval}
\label{subsec:ir}

Information Retrieval (IR) for Indian languages spans monolingual, cross-lingual, mixed-script, spoken-query, ontology-driven, and document-structure–aware settings. Foundational work includes early CLIR systems \cite{jagarlamudi2007cross} and synset-based Telugu IR \cite{ramakrishna2013information}. Recent multilingual benchmarks include IndicIRS-Suite \cite{haq2024indicirsuite} and Hindi-BEIR \cite{acharya2024hindi}. Cross-lingual advances span word-vector community methods \cite{bhattacharya2018using}, mixed-script query expansion \cite{gupta2014query}, spoken-query CLIR via SqCLIRIL \cite{dave2025sqcliril}, and benchmarks such as Anveshana for English–Sanskrit retrieval \cite{jagadeeshan2025anveshana}. Low-resource IR is supported through Hindi optimization strategies \cite{sourabh2012query}, massively multilingual fact-extraction models \cite{singh2022massively}, and Urdu resources including CURE \cite{iqbal2021cure} and earlier IE-based systems \cite{mukund2010information}. Domain-specific and structured retrieval includes Tamil ontology-based IR \cite{sankaralingam2017onto} and spatially aware document extraction via IndicCharGrid \cite{trivedi2025indicchargrid}. Detailed resources and benchmarks are provided in Appendix (Table~\ref{tab:indic_ir_survey} and Figure~\ref{fig:ir}).

\subsection{Dialogue Systems}
\label{subsec:dialogue}

Dialogue systems for Indian languages span task-oriented, open-domain, multilingual, and code-mixed settings. Early resources include code-mixed goal-oriented datasets \cite{banerjee2018dataset}, Dravidian task-oriented systems \cite{kanakagiri2021task}, and large-scale conversational subtitles via IndicDialogue \cite{arnob2024indicdialogue}. Task-oriented datasets further include Hindi dialogue state tracking \cite{malviya2021hdrs}, TamilATIS for intent and slot filling \cite{ramaneswaran2022tamilatis}, hope-speech dialogue data \cite{chakravarthi2020hopeedi}, and multilingual transport-domain dialogs \cite{ambastha2021mtransdial}. Broader multilingual and dialectal coverage is supported by MD3 \cite{eisenstein2023md3} and chat-translation benchmarks \cite{gain2022low}.

Applied dialogue systems increasingly target real-world use cases, including healthcare \cite{badlani2021multilingual,singh2023ai}, rural and agricultural assistance \cite{mehra2025empowering,anand2023multi}, COVID-19 support \cite{thara2024multilingual}, and regional-language services for Odia \cite{agarwal2023generative}, Assamese \cite{sarma2023shiksha}, and Urdu \cite{mohiuddin2023multilingual}. Additional research explores multilingual chatbot architectures \cite{singh2023multilingual} and personality modeling in Hindi conversations \cite{kumar2024hindipersonalitynet}. Further details are reported in Appendix (Table~\ref{tab:indic_dialogue_systems} and Figure~\ref{fig:DIALOGUE}).

\section{Speech and Multimodality}

\subsection{Speech Technologies}
\label{subsec:speech}

Indian-language speech research spans Automatic Speech Recognition (ASR), Text-to-Speech (TTS), speech translation, language/accent identification, and dataset creation. Core multilingual resources include IndicSUPERB \cite{javed2023indicsuperb}, IndicVoices/IndicVoices-R \cite{javed2024indicvoices,sankar2024indicvoices}, IndicSpeech \cite{srivastava2020indicspeech}, LDC-IL \cite{choudhary2020ldc}, IIITH-ILSC \cite{vuddagiri2018iiith}, regional and endangered-language corpora \cite{basu2021multilingual,kumar2023collecting}, dialect datasets \cite{podila2022telugu}, and Hinglish speech \cite{ganji2019iitg}. ASR work covers accented and low-resource benchmarks \cite{javed2024lahaja,rakib2023ood,londhe2018chhattisgarhi,anoop2023suitability,shetty2021exploring}, large corpora \cite{bhogale2023vistaar,sharma2023hindispeech,kalluri2021nisp,ahamad2020accentdb,vancha2022word}, and toolkits such as Vakyansh \cite{chadha2022vakyansh}. TTS advances include multilingual and expressive systems \cite{prakash2019building,he2020open,varadhan2024rasa,sathiyamoorthy2024unified,sharma2025indicsynth}. Speech translation resources span Bhasaanuvaad and large ST corpora \cite{jain2024bhasaanuvaad,sankar2025towards,sethiya2025indic,sethiya2024indic,shah2025indicst}, while language/accent ID and special domains include multimodal LID \cite{puthran2025multimodal}, northeastern ID \cite{basu2021multilingual}, clinical speech \cite{vekkot2023dementia}, and speech-to-intent datasets \cite{rajaa2022skit}. Further details are provided in Appendix (Table~\ref{tab:indic_speech_resources} and Figure~\ref{fig:speech}).

\subsection{Multimodal Language Understanding}
\label{subsec:multimodal}

Multimodal NLP for Indian languages spans vision--language grounding, OCR, scene text, handwriting, and document understanding. Representative multilingual datasets include Chitrakshara \cite{khan2025chitrakshara}, Bengali and Hindi Visual Genome variants \cite{sen2022bengali,parida2019hindi}, Dravidian multimodal MT \cite{chakravarthi2019multilingual}, conversational resources such as M2H2 \cite{chauhan2021m2h2}, document-level VLM pretraining via Nayana \cite{kolavi2025nayana}, and sign-language understanding through INCLUDE \cite{sridhar2020include}. OCR and scene-text research covers multilingual Indic OCR \cite{mathew2016multilingual}, synthetic benchmarks \cite{saini2022ocr}, post-OCR Sanskrit correction \cite{maheshwari2022benchmark}, script-specific systems for Kannada \cite{kumar2020lipi}, Tamil handwriting \cite{shaffi2021uthcd}, Gujarati character recognition \cite{pareek2020gujarati}, and large scene-text datasets such as Bharat Scene Text \cite{de2025bharat} and IndicSTR12 \cite{lunia2023indicstr12}. Handwriting recognition and script identification are supported by iiit-indic-hw-words \cite{gongidi2021iiit}, Gurmukhi stroke datasets \cite{singh2016online}, multi-script handwritten corpora \cite{alaei2012dataset}, Kannada-MNIST \cite{prabhu2019kannada}, Kannada document scans \cite{alaei2011benchmark}, PHDIndic\_11 \cite{obaidullah2018phdindic_11}, Bengali grapheme datasets \cite{alam2021large}, and mixed-script document benchmarks \cite{singh2018benchmark}. Full details are provided in Appendix (Table~\ref{tab:multimodal_indic_resources}, Figure~\ref{fig:multimodal}).

\section{Societal, Cultural, and Emerging Tasks}

\subsection{Misinformation and Fact Checking}

Indic misinformation research spans fake news detection, fact-check repositories, and multimodal and multilingual modeling. Regional datasets cover Tamil \cite{mirnalinee2022novel}, Malayalam \cite{devika2024dataset,sujan2023malfake}, Manipuri (Romanized) \cite{devi2025dataset}, and Urdu \cite{amjad2020bend}, alongside multilingual resources for Tamil--Malayalam \cite{hariharan2022impact}, mixed Indic languages \cite{sivanaiah2022fake,raja2023fake}, and Hindi--Marathi--Telugu \cite{thaokar2022multi}. Hindi-centric resources include LFWE \cite{sharma2023lfwe}, IFND \cite{sharma2023ifnd}, Hindi fake-news corpora \cite{kumar2025fake}, and systems such as DeFactoX \cite{bansal2025fragments}, Aletheia \cite{badam2022aletheia}, while large-scale repositories include FakeNewsIndia \cite{dhawan2022fakenewsindia}, FactDrill \cite{singhal2022factdrill}, BharatFakeNewsKosh \cite{singh2023bharatfakenewskosh}, and COVID-related datasets \cite{kar2021no,shahi2020fakecovid}. Multimodal resources include Hindi affect-enriched data \cite{kumar2025sifting}, Tamil multimodal fact-checking \cite{francis2024tamilfacts,francis2025multimodal}, multilingual multimodal models such as MMCFND \cite{bansal2024mmcfnd}, MALFake \cite{sujan2023malfake}, Indian deepfake datasets \cite{das2025indeepfake}, and MMM \cite{gupta2022mmm}, with claim detection and verification supported by fact-check factorization \cite{singhal2021factorization}, Twitter claim identification \cite{dutta2022multilingual}, and the FakeRealIndian benchmark \cite{tufchi2023fakerealindian}. Details are deferred to Appendix (Table~\ref{tab:indic_misinformation}, Figure~\ref{fig:misinfo}).

\subsection{Cultural Knowledge \& Understanding}

Cultural NLP research spans textual, multimodal, and cross-cultural evaluation. Broad resources include PARIKSHA \cite{watts2024pariksha}, D3CODE \cite{davani2024d3code}, D-PLACE \cite{kirby2016d}, Global Jukebox \cite{wood2022global}, and culturally aware NLI \cite{huang2023culturally}. Indian-focused studies examine mental health expressions \cite{rai2025cross}, subcultural and traditional knowledge \cite{chhikara2025through}, stigma \cite{jonnala2025decoding}, tourism QA \cite{gatla2025tourism}, indigenous food \cite{gogoi2025s}, Indian art music \cite{srinivasamurthy2021saraga}, poetry \cite{jamil2026crossing}, idioms \cite{das2026meaning} and social artifacts \cite{seth2024dosa}. Large-scale Indic benchmarks include DRISHTIKON \cite{maji2025drishtikon}, SANSKRITI \cite{maji2025sanskriti}, DIWALI \cite{sahoo2025diwali}, VIRAASAT \cite{surana2026viraasat}, Pragyaan \cite{rachamalla2025pragyaan}, and the benchmark suite in \cite{doddapaneni2023towards}.

Cultural alignment and evaluation of LLMs is advanced through NativQA \cite{hasan2025nativqa}, CulturePark \cite{li2024culturepark}, CultureLLM \cite{li2024culturellm}, culturally sensitive analyses \cite{banerjee2025navigating}, and multilingual foundations such as Krutrim LLM \cite{kallappa2025krutrim}. Figurative language understanding is explored in \cite{kabra2023multi}. Multimodal cultural understanding is supported by CultureVLM \cite{liu2025culturevlm}, Bhasa \cite{leong2023bhasa}, Multi$^{3}$Hate \cite{bui2025multi3hate}, VLM cultural benchmarks \cite{nayak2024benchmarking}, and affective multimedia datasets such as AFDI \cite{mishra2023affective}. Furthermore, a significant portion of Indian-language resources relies on translation-based construction pipelines, which enable rapid scaling but may fail to capture indigenous linguistic, pragmatic, and socio-cultural nuances. This introduces a trade-off between scalability and cultural fidelity, highlighting the need for more native, community-driven data collection efforts. A detailed comparison is provided in Appendix (Table~\ref{tab:cultural_nlp} and Figure~\ref{fig:cultural}).

\subsection{Emerging Directions}

Indian NLP is rapidly expanding across bias evaluation, code-mixing, style transfer, and domain-specific reasoning. Bias and fairness research introduces India-centric benchmarks such as IndiBias \cite{sahoo2024indibias}, Indian-BHeD \cite{khandelwal2024indian}, and DBNLP \cite{tg2025dbnlp}, alongside extensive embedding and LM bias analyses across social, cultural, caste, and gender dimensions \cite{bansal2021debiasing,tiwari2022casteism,malik2022socially,das2023toward,vashishtha2023evaluating,hada2024akal,sahoo2023prejudice,ghate2024evaluating,kumar2022comma,mukherjee2023global,santhosh2025indicasa,khurana2022animojity,kumar2024multilingual,joshi2024since,pujari2019debiasing,aneja2025beyond,khadilkar2022gender,kamruzzaman2025banstereoset}. Code-mixing research spans NER, sentiment, offensive language, and language identification across Hinglish, Bangla--English--Hindi, Kannada--English, Sinhala--English, and Gujarati--Hindi \cite{priyadharshini2020named,goswami2023offmix,nayak2022l3cube,hande2020kancmd,smith2019language,kazi2020sentence,maity2021multi,bali2014borrowing,bhargava2016named,sheth2025comi,dey2024bharatbhasanet,chatwal2024overcoming,kodali2022symcom,sandhan2022prabhupadavani}. Multilingual style transfer and controllable generation are advanced through new datasets and low-resource or few-shot models \cite{mukherjee2024multilingual,krishna2022few,mukherjee2023text,gunna2021transfer,mukherjee2023low,nag2023transfer,kumar2019augmented,protasov2025evaluating,ghosal-etal-2025-relic}. Domain-specific reasoning benchmarks evaluate mathematical, legal, cultural, analogical, and scientific capabilities of LLMs \cite{anand2025multilingual,nigam2025nyayaanumana,joshi2024since,singh2025indic,gupta2025hats,bandooni2025ganitbench,methani2020plotqa,mukherjee2025mmjee,acharya2020towards,saxena2025riddle,onyame2026cure,ghosh2025clinic,ghosh2025survey}, complemented by broader multilingual evaluations covering generalization, retrieval-augmented generation, and code generation \cite{singh2024indicgenbench,joshi2025elr,khan2025quench,singh2025indiceval,khan2025chitrarth,prasanjith2025indicragsuite,maheshwari2025parambench}. Detailed resources are deferred to the Appendix (Tables~\ref{tab:bias_indic}, \ref{tab:codemix_indic}, \ref{tab:style_indic}, \ref{tab:reasoning_indic} and Figure~\ref{fig:newtopics}).

\section{Future Directions}
While this survey consolidates existing resources and benchmarks, several open challenges remain in scaling Indian-language NLP toward equitable and culturally grounded systems. Key directions include expanding coverage beyond high-resource languages, improving fine-grained and context-aware evaluation, and addressing societal considerations such as bias, code-mixing, and responsible deployment. We provide a detailed discussion of these challenges and research opportunities in Appendix~\ref{sec:future_directions}. Additionally, a detailed discussion of cross-cutting gaps is deferred to Appendix~\ref{sec:unified_gaps}. 

\paragraph{Toward Standardized Evaluation Reporting.}
A recurring challenge across Indian NLP resources is inconsistent and incomplete evaluation reporting. To improve reproducibility and comparability, we highlight a minimal set of recommended reporting practices: 
(i) clear train/dev/test splits, 
(ii) explicit metric definitions and justification, 
(iii) documentation of annotation procedures and inter-annotator agreement (where applicable), 
(iv) specification of language, dialect, and script coverage, and 
(v) domain and data collection context. 
We emphasize that these recommendations are illustrative rather than prescriptive, but can serve as a starting point toward more standardized evaluation practices.

% \paragraph{Resource-Efficient Modeling and Accessibility.}
% While large-scale pretrained models increasingly dominate Indian NLP, their computational and infrastructure requirements may limit accessibility in low-resource and regionally constrained settings. A detailed analysis of efficiency and hardware considerations is beyond the scope of this survey, which focuses on datasets, benchmarks, and task-level modeling trends. Nevertheless, this highlights an important direction for future work, including resource-efficient modeling, lightweight adaptation techniques, and infrastructure-aware deployment strategies tailored to diverse contexts.

\paragraph{Resource-Efficient Modeling and Accessibility.}
Large-scale pretrained models dominate Indian NLP but may be inaccessible in low-resource settings due to computational demands. A detailed analysis of efficiency and hardware is beyond the scope of this survey; however, resource-efficient modeling and lightweight deployment remain important directions for future work.

\section{Conclusion}

This survey provides a unified, task-centric overview of Indian NLP resources across text, speech, multimodality, and societal and cultural dimensions. By organizing a fragmented literature into a coherent taxonomy, we highlight both substantial progress and persistent gaps in language coverage, annotation practices, and evaluation. Overall, the landscape reflects a shift toward multilingual, multimodal, and culturally grounded modeling, alongside a continued need for coordinated and inclusive research efforts.

\section*{Limitations}

Despite broad coverage, this survey cannot fully capture the rapidly evolving Indian NLP ecosystem, where new datasets, models, and evaluations emerge continuously. Our synthesis is based on reported findings rather than reproduced experiments, and certain areas, such as endangered languages, conversational and dialectal speech, handwriting OCR, and multi-script multimodal tasks, remain underrepresented due to limited publicly available resources. Space constraints also preclude detailed analysis of implementation choices and architectural variations, and some industry or poorly documented datasets may be absent. Nevertheless, the survey aims to provide the most comprehensive and structured snapshot currently feasible of Indian NLP research.

\section*{Ethical considerations}

This survey synthesizes publicly available datasets, benchmarks, and models without introducing new data; however, the reviewed resources raise important ethical concerns. Indian-language datasets often encode culturally sensitive attributes (e.g., caste, gender, religion, region), and tasks such as hate speech detection, misinformation analysis, cultural reasoning, and bias evaluation inherently engage with socio-political content. Many corpora rely on scraped social media data without explicit consent, raising privacy, consent, and data-provenance issues, while annotation processes may reflect cultural subjectivity or encode harmful stereotypes. Cross-lingual and multimodal resources may conflate dialects, marginalize communities, or flatten cultural nuance, and models trained on such data risk propagating or amplifying systemic biases, especially in generative and high-impact settings. We emphasize the need for transparent documentation, culturally informed data practices, inclusive collection protocols, and rigorous bias and safety evaluations, and encourage future work to explicitly address these challenges.

\section*{Acknowledgements}

Raghvendra Kumar gratefully acknowledges the support of the Prime Minister’s Research Fellowship (PMRF) in carrying out this research.

\bibliography{custom}

\appendix
\section{Task Definitions}
\label{definitions}

This appendix provides formal definitions and contextual descriptions of the seventeen NLP tasks considered in this survey. These definitions complement the main text and are intended to offer precise task-level grounding without interrupting the high-level narrative.

\subsection{Core Linguistic Processing}

\textbf{Tokenization, normalization, and morphological analysis} concern the earliest stages of text processing, where raw textual input is segmented into basic units, standardized across orthographic variations, and analyzed for morphological structure. These steps are particularly challenging for Indian languages due to rich inflection, compounding, script variation, and spelling diversity \cite{mielke2021between,huang2023normalization,antony2012computational}.

\textbf{Part-of-speech (POS) tagging} involves assigning syntactic category labels (e.g., noun, verb, adjective) to each token in a sentence. Accurate POS tagging is foundational for higher-level syntactic and semantic tasks and is complicated in Indian languages by free word order and morphological richness \cite{chiche2022part,antony2011parts,rathod2015survey}.

\textbf{Named entity recognition (NER)} focuses on identifying and classifying mentions of real-world entities such as persons, locations, organizations, and dates within text. Indian-language NER must account for sparse capitalization cues, transliteration variability, and limited annotated resources \cite{li2020survey,bhattacharjee2019named,patil2016survey}.

\subsection{Text Classification and Semantics}

\textbf{Sentiment and emotion analysis} aim to detect subjective polarity (e.g., positive, negative, neutral) and affective states expressed in text. These tasks are widely applied to social media, reviews, and public discourse, and pose additional challenges in Indian languages due to code-mixing and cultural expression of emotions \cite{nandwani2021review,wankhade2022survey,yadollahi2017current}.

\textbf{Hate speech and toxicity detection} involve identifying abusive, harmful, or discriminatory language targeting individuals or groups. This task is socially critical and technically challenging in the Indian context due to linguistic diversity, implicit abuse, and socio-political nuance \cite{fortuna2018survey,chhabra2023literature,nandi2024survey}.

\textbf{Topic classification} assigns documents or text segments to predefined thematic categories, enabling content organization, filtering, and retrieval. Indian-language topic classification often suffers from limited labeled data and domain imbalance \cite{wu2024survey,qiang2020short}.

\textbf{Natural language understanding (NLU)} encompasses a broad set of semantic tasks, including textual inference, paraphrase detection, and semantic similarity estimation. These tasks test a model’s ability to capture meaning beyond surface forms and remain underexplored for many Indian languages \cite{storks2019recent,zhou2021paraphrase,chandrasekaran2021evolution}.

\subsection{Generation and Translation}

\textbf{Summarization} seeks to generate concise representations of longer texts while preserving key information. Indian-language summarization spans extractive and abstractive paradigms and is applied across domains such as news, legal documents, and social media \cite{el2021automatic,mamidala2021text,shimpikar2017survey}.

\textbf{Machine translation} focuses on automatically translating text between languages. For Indian languages, challenges include morphological divergence, syntactic variation, and limited parallel corpora, especially for low-resource language pairs \cite{dabre2020survey,stahlberg2020neural,lopez2008statistical}.

\textbf{Question answering (QA)} involves extracting or generating answers to natural language queries based on a given context or knowledge source. Indian-language QA remains constrained by dataset scarcity and linguistic diversity across question formulations \cite{ojokoh2018review,kolhatkar2023indic}.

\subsection{Retrieval and Interaction}

\textbf{Information retrieval (IR)} addresses the task of locating and ranking relevant documents or passages in response to a user query. Cross-lingual and multilingual IR are especially relevant in the Indian setting, where queries and documents may appear in different languages or scripts \cite{hambarde2023information,bajpai2014cross}.

\textbf{Dialogue systems} model interactive conversational agents capable of maintaining context and generating appropriate responses. Indian-language dialogue systems face challenges related to code-mixing, informal speech, and limited conversational datasets \cite{motger2022software,gao2018neural}.

\subsection{Speech and Multimodality}

\textbf{Speech processing} includes automatic speech recognition (ASR) and text-to-speech (TTS) synthesis for spoken language. Indian languages exhibit wide phonetic diversity, accent variation, and resource imbalance, making robust speech modeling particularly challenging \cite{bhangale2022survey,panda2020survey}.

\textbf{Multimodal language understanding} integrates textual information with visual signals such as images, documents, or OCR-extracted text. This task is increasingly important for real-world Indian applications involving scanned documents, social media, and low-literacy settings \cite{nguyen2021survey,xu2023multimodal}.

\subsection{Societal, Cultural, and Emerging Tasks}

\textbf{Misinformation and fact-checking} aim to detect false, misleading, or manipulated content across text, images, and videos. The multilingual and multimodal nature of misinformation in India poses significant challenges for automated verification systems \cite{zhou2020survey,sivanaiah2022fake}.

\textbf{Cultural reasoning} focuses on modeling culturally grounded knowledge, norms, and inferences that are specific to particular communities or regions. Such reasoning is essential for fair and context-aware NLP systems in culturally diverse societies like India \cite{liu2025culturally,pawar2025survey}.

\textbf{Emerging tasks} encompass a range of developing research areas, including code-mixing analysis, bias and fairness assessment, domain-specific reasoning, and stylistic text transformation. These tasks reflect evolving societal needs and highlight open challenges for inclusive and responsible NLP \cite{winata2023decades,blodgett2020language,yu2024natural,jin2022deep}.

\section{Unified Gaps Across the Indian NLP Landscape}
\label{sec:unified_gaps}

Despite significant progress across tasks and modalities, Indian NLP continues to face several systemic and cross-cutting challenges that limit scalability, generalization, and real-world impact. \textbf{Data coverage remains highly uneven}: a small set of relatively high-resource languages (e.g., Hindi, Tamil, Telugu, Bengali, Marathi) benefit from multiple datasets across tasks, while most dialects, tribal and endangered languages, and smaller Indic varieties lack large-scale, clean, and domain-diverse corpora. This imbalance constrains reliable evaluation, weakens cross-lingual transfer, and limits effective adaptation of large language models (LLMs) to low-resource settings.

\textbf{Annotation practices are inconsistent and fragmented across tasks}. Many datasets employ divergent label taxonomies, task-specific heuristics, or ad hoc annotation guidelines, often with limited documentation of annotator training, agreement statistics, or quality-control procedures. This reduces interoperability across datasets for tasks such as NER, sentiment and emotion analysis, hate speech detection, OCR, speech processing, and cultural reasoning, and hinders unified benchmarking or model reuse. In several domains, annotations are also narrowly scoped to surface-level phenomena, overlooking pragmatic, discourse-level, or culturally grounded aspects of language use.

\textbf{Task and domain coverage remains skewed}. Social-media data dominates sentiment, hate speech, and misinformation research, while summarization is disproportionately centered on news and legal text. Multimodal and speech datasets cover only a limited range of scripts, dialects, acoustic conditions, and visual styles, restricting robustness across regions and user populations. Core linguistic challenges unique to the Indian context, such as pervasive code-mixing, romanization, spelling variation, and non-standard orthographies, remain insufficiently addressed across foundational tasks, including tokenization, tagging, NLU, MT, QA, and IR.

\textbf{Evaluation protocols lack standardization}. Metrics, difficulty settings, and train--test splits vary widely across languages and modalities, with few shared leaderboards or reproducible benchmarks, particularly for multimodal, OCR, speech, and LLM-centric tasks. As a result, cross-task and cross-language comparisons are often unreliable. Moreover, \textbf{cross-lingual and cross-modal generalization remains weak}, with many models exhibiting sharp performance drops outside the languages, domains, or modalities seen during training.

Finally, \textbf{India-specific bias, safety, and cultural alignment remain under-evaluated}. While recent work has begun to explore bias along dimensions such as caste, religion, gender, region, and sociolect, systematic evaluation is still limited—especially in multimodal, generative, and long-form reasoning settings. These gaps collectively highlight the need for scalable, standardized, and culturally grounded datasets, benchmarks, and modeling strategies to support inclusive, robust, and socially responsible Indian-language NLP.

\section{Frequently Asked Questions (FAQs)}\label{sec:faq}

\begin{enumerate}

\item \textbf{What qualifies a resource to be included in this survey?}  
We include datasets, benchmarks, and tools developed specifically for Indian languages, as well as multilingual resources that explicitly cover Indian languages (including English--Indic settings).

\item \textbf{Why are some languages grouped under the ``Indic Languages'' category in figures?}  
Resources covering multiple Indian languages (often 15--200) are aggregated under the \emph{Indic Languages} category, while resources focused exclusively on a single language are counted toward that language.

\item \textbf{Does the survey aim to be exhaustive or representative?}  
The survey prioritizes breadth and diversity over completeness, selecting representative resources to reflect methodological trends, task coverage, and language diversity rather than listing every available work.

\item \textbf{Why is English included in some datasets discussed in the survey?}  
English is included when it appears alongside Indian languages in multilingual or code-mixed resources, as such settings are common in real-world Indian NLP applications.

\item \textbf{How does this survey differ from existing Indic or multilingual NLP surveys?}  
Unlike prior surveys that focus on specific tasks or embed Indian languages within broader multilingual contexts, this work provides a unified, task-centric view dedicated exclusively to Indian NLP.

\item \textbf{Why are certain tasks (e.g., sentiment, hate speech) more resource-rich than others?}  
These tasks often rely on easily available social-media data, whereas tasks such as multimodal reasoning, speech processing, and low-resource language modeling require more complex and costly data collection.

\item \textbf{Are pretrained LLMs and foundation models fully solving Indian NLP challenges?}  
While multilingual pretrained models have improved coverage, significant gaps remain in low-resource languages, cultural grounding, bias mitigation, and cross-modal generalization.

\item \textbf{How are annotation quality and consistency addressed in the survey?}  
We highlight annotation practices, agreement reporting, and documentation where available, and identify inconsistent labeling and sparse metadata as key cross-cutting challenges.

\item \textbf{Why is code-mixing treated as a recurring challenge across tasks?}  
Code-mixing and romanization are pervasive in Indian language use and affect nearly all NLP pipelines, from tokenization to generation, making them foundational rather than task-specific issues.

\item \textbf{What are the main limitations of current evaluation practices?}  
Evaluation protocols vary widely across languages and tasks, with inconsistent metrics, difficulty levels, and benchmarks, limiting reliable cross-language and cross-task comparison.

\item \textbf{How does the survey address societal and cultural dimensions of NLP?}  
Dedicated sections cover misinformation, cultural reasoning, bias, and emerging tasks, emphasizing India-specific social, cultural, and ethical considerations often overlooked in generic NLP surveys.

\item \textbf{Where can readers find detailed tables and extended comparisons?}  
Comprehensive task-wise tables, language-wise distributions, and unified gap analyses are provided in the appendix and referenced throughout the paper.

\item \textbf{Who is this survey intended for?}  
The survey is intended for NLP researchers, dataset creators, model developers, practitioners, and policymakers interested in building inclusive, culturally grounded AI for Indian languages.

\end{enumerate}

\section{Future Directions}
\label{sec:future_directions}

Despite rapid progress across datasets, benchmarks, and models, Indian-language NLP continues to face distinctive challenges arising from linguistic diversity, uneven resource availability, and socio-cultural complexity. Based on the trends and gaps identified throughout this survey, we outline several future directions that can guide sustained and equitable development of NLP for Indian languages.

\subsection{Balanced Language Coverage Beyond High-Resource Languages}
A consistent pattern across nearly all tasks is the concentration of resources around a small set of high-resource languages, while many scheduled and non-scheduled languages remain underrepresented. Future research should prioritize expanding coverage to low-resource, endangered, and regionally marginalized languages, not only in basic classification tasks but also in higher-level reasoning, generation, and interaction. Scalable approaches such as cross-lingual transfer, typology-aware modeling, multilingual pretraining, and community-driven data collection offer promising pathways but require broader adoption and systematic evaluation.

\subsection{Evaluation Beyond Aggregate Metrics}
Current evaluation practices frequently rely on aggregate scores that mask disparities across languages, scripts, dialects, and domains. Future benchmarks should emphasize disaggregated evaluation, reporting performance across language families, script variations, code-mixed settings, and sociolinguistic contexts. Task-specific evaluation protocols that capture robustness to spelling variation, colloquial usage, and domain shift are particularly important for realistic assessment of Indian-language NLP systems.

\subsection{Culturally Grounded and Context-Aware Modeling}
While recent work has begun to address cultural knowledge and reasoning, most NLP systems still struggle with localized cultural context, implicit norms, and region-specific associations. Future research should move beyond surface-level factual recall toward deeper cultural understanding, integrating historical, social, and regional context into both modeling and evaluation. This direction calls for culturally grounded annotation guidelines, interdisciplinary collaboration, and benchmarks that explicitly test contextual and culturally situated reasoning.

\subsection{Responsible and Inclusive NLP for Societal Applications}
Many Indian-language NLP applications operate in socially and politically sensitive settings, including misinformation detection, hate speech moderation, and public-service dialogue systems. Future work should place greater emphasis on responsible and inclusive design, addressing annotation bias, dataset provenance, and representational harms. Participatory data creation, transparent documentation, and evaluation frameworks sensitive to caste, gender, religion, and regional identity are essential to ensure that NLP systems do not reinforce existing social inequities.

\subsection{Code-Mixing as a First-Class Phenomenon}
Code-mixing is pervasive across Indian languages but is still often treated as a secondary or noisy setting. Future research should treat code-mixing as a first-class linguistic phenomenon, with dedicated datasets, annotation schemes, and modeling approaches that reflect realistic language use. This includes multi-script code-mixing, spoken code-mixed data, and task diversity beyond sentiment and hate speech, such as summarization, question answering, and dialogue.

\subsection{Scaling Multimodal and Speech Resources Equitably}
Although multimodal and speech-based NLP has seen significant growth, coverage remains uneven across languages, domains, and dialects. Future directions include expanding speech, vision--language, and document understanding resources for low-resource languages, improving accent and dialect diversity, and grounding multimodal benchmarks in Indian contexts such as education, healthcare, and governance. Resource-efficient multimodal modeling and unified evaluation across text, speech, and vision remain open challenges.

\subsection{Bridging Research and Deployment}
Many surveyed resources are developed in academic settings with limited attention to deployment, maintenance, and long-term usability. Future work should emphasize end-to-end evaluation, including robustness, interpretability, and failure analysis in real-world systems. Open-source tooling, standardized documentation, and long-term dataset stewardship are crucial for translating research advances into sustainable and impactful applications.

\subsection{Unified Benchmarking and Longitudinal Evaluation}
The rapid proliferation of datasets and benchmarks has led to fragmentation and limited comparability across studies. A promising future direction is the development of unified and extensible benchmark suites that support longitudinal evaluation across tasks, languages, and model families. Such benchmarks can facilitate principled comparisons, track progress over time, and help identify persistent gaps in language coverage and model capabilities.

\paragraph{Summary}
Overall, future progress in Indian-language NLP will depend not only on scaling data and models, but also on balanced language coverage, culturally grounded evaluation, responsible system design, and sustained community engagement. Addressing these challenges is essential for building inclusive, trustworthy, and socially meaningful NLP systems that reflect the full linguistic and cultural diversity of India.

\section{Resource Collection and Screening Methodology}
\label{sec:resource_collection}

\subsection{Data Sources and Retrieval Strategy}
To construct a comprehensive and representative inventory of Indian NLP resources, we systematically curated publications and artifacts from a diverse set of sources. These include major NLP and speech venues such as ACL, EMNLP, NAACL, COLING, LREC, and Interspeech, as well as preprint repositories like arXiv. In addition, we incorporated resources from institutional and community-driven repositories, including AI4Bharat, LDC-IL, and the IIIT Hyderabad Language Sciences Center (IIITH-ILSC), which host several datasets and benchmarks not formally published in conferences.

The initial pool of candidate resources was identified using task-specific and language-specific keyword queries (e.g., ``Indic NLP'', ``Hindi dataset'', ``code-mixed corpus'', ``low-resource MT'', ``ASR Indian languages''), combined with targeted searches for known benchmark suites such as IndicNLPSuite and IndicLLMSuite. We further expanded this pool using citation chaining, exploring both forward and backward citations of influential works to ensure coverage of foundational as well as recent contributions.

\subsection{Inclusion Criteria}
Resources were included based on the following criteria:
\begin{itemize}
    \item \textbf{Relevance to Indian Languages:} The resource must focus on one or more Indian languages, including both high-resource (e.g., Hindi, Tamil) and low-resource or endangered languages.
    \item \textbf{Task Alignment:} The resource must correspond to a well-defined NLP, speech, or multimodal task (e.g., machine translation, text classification, ASR, information extraction, vision-language tasks).
    \item \textbf{Availability:} Preference was given to publicly available datasets, benchmarks, or documented resources. However, widely cited but restricted-access datasets (e.g., LDC-IL collections) were also included for completeness.
    \item \textbf{Scholarly or Practical Impact:} Resources introduced or used in peer-reviewed publications, benchmark papers, or widely adopted toolkits were prioritized.
\end{itemize}

\subsection{Exclusion Criteria}
We excluded:
\begin{itemize}
    \item Works that do not introduce, utilize, or evaluate datasets or corpora.
    \item Duplicated datasets reported across multiple papers without meaningful modifications.
\end{itemize}

\subsection{Screening and Deduplication Process}
The collected resources were subjected to a multi-stage screening process. First, titles and abstracts were manually reviewed to filter out irrelevant entries. Next, full-text inspection was performed to verify task definitions, dataset characteristics, and language coverage. 

To address redundancy, we performed deduplication by identifying datasets that appeared across multiple publications (e.g., benchmark suites reused in subsequent works). In such cases, we retained the original source paper while noting derivative usages where relevant.

\subsection{Metadata Extraction and Annotation}
For each selected work, we performed structured metadata extraction aligned with the survey taxonomy and tabular schema. Specifically, each paper was annotated along the following dimensions:

\begin{itemize}
    \item \textbf{Paper:} Bibliographic information including authors, venue, and year, serving as the primary unit of analysis.
    
    \item \textbf{Focus \& Objective:} The primary research goal of the work, such as dataset creation, benchmark development, model proposal, or task-specific evaluation.
    
    \item \textbf{Languages:} The set of Indian languages covered, including both single-language and multilingual settings, with explicit identification of code-mixed or cross-lingual scenarios where applicable.
    
    \item \textbf{Methodologies \& Algorithms:} The key techniques employed, including model architectures (e.g., transformer-based models), training strategies (e.g., pretraining, fine-tuning), and evaluation protocols.
    
    \item \textbf{Key Resources / Contributions:} The datasets, benchmarks, tools, or corpora introduced or utilized, along with their characteristics (e.g., scale, modality, task coverage).
    
    \item \textbf{Key Findings:} The main empirical or conceptual insights reported, including performance trends, challenges, and limitations identified in the context of Indian languages.
\end{itemize}

All annotations were performed through careful full-text review to ensure consistency and fidelity to the original contributions. This structured representation enables systematic comparison across works and supports downstream analysis of trends across tasks, languages, and methodologies.

This metadata was used to construct a unified taxonomy spanning 17 task categories and multiple modalities, enabling fine-grained analysis of trends across languages and tasks.

\subsection{Quality Control and Coverage Validation}
To ensure robustness and coverage, we cross-validated our collection against existing surveys and benchmark compilations (e.g., IndicNLPSuite, IndicLLMSuite, and language-specific surveys such as Marathi NLP). Any missing but relevant resources identified through this comparison were incorporated iteratively.

Finally, we performed consistency checks across annotations and resolved discrepancies through manual inspection, ensuring uniform categorization across all resources included in the survey.

\section{Licensing and Usage Constraints}
\label{sec:licensing}

We examined licensing and usage information for the surveyed resources where available. However, explicit licensing details are inconsistently reported across the literature, particularly for earlier datasets and model-centric studies. Consequently, a substantial portion of resources lack clearly documented licensing terms in their original publications or repositories.

Given this variability, we do not provide exhaustive licensing tables. Instead, we report licensing information where explicitly stated (e.g., open-source or research-only usage) and indicate when such details are unavailable. Overall, this reflects a broader ecosystem-level gap, highlighting the need for standardized and transparent reporting of licensing and usage conditions in Indian NLP resources.

\paragraph{Tokenization, Normalization, and Morphological Analysis.}
Licensing information is available for only a small subset of resources. The iNLTK toolkit \cite{arora2020inltk} is released as an open-source library. Several datasets, including the word similarity dataset \cite{akhtar2017word}, GujMORPH \cite{baxi2022gujmorph}, the Sanskrit segmentation dataset \cite{krishna2017dataset}, and the multiword expressions dataset \cite{singh2016multiword}, are distributed for research use, although formal licenses are often undocumented.

\paragraph{Part-of-Speech Tagging.}
POS tagging resources generally lack explicit licensing documentation. Datasets such as the Bengali news corpus \cite{ekbal2008web} and the Magahi POS dataset \cite{kumar2012developing} are primarily used in academic settings, with implied research-only usage but without clearly specified licenses.

\paragraph{Named Entity Recognition.}
Several NER datasets, including HiNER \cite{murthy2022hiner}, Naamapadam \cite{mhaske2023naamapadam}, L3Cube-MahaNER \cite{litake2022l3cube}, B-NER \cite{haque2023b}, and AsNER \cite{pathak2022asner}, are publicly released for benchmarking and research. However, explicit licensing terms are not consistently documented across these resources.

\paragraph{Sentiment and Emotion Analysis.}
Most sentiment and emotion datasets are released as research benchmarks with implicit academic usage. Examples include IndiSentiment140 \cite{kumar2024indisentiment140}, DravidianCodeMix \cite{chakravarthi2022dravidiancodemix}, L3Cube-MahaSent \cite{kulkarni2021l3cubemahasent}, HindiMD \cite{ekbal2022hindimd}, EmoInHindi \cite{singh2022emoinhindi}, and SentMix-3L \cite{raihan2023sentmix}. Emotion-focused datasets such as Anubhuti \cite{pal2020anubhuti}, Bhaav \cite{kumar2019bhaav}, K{\=a}vi \cite{saini2020kavi}, and TamilEmo \cite{vasantharajan2022tamilemo} follow similar patterns, with limited formal licensing documentation.

\paragraph{Hate Speech and Toxicity Detection.}
Datasets such as HopeEDI \cite{chakravarthi2020hopeedi}, TABHATE \cite{sharma2024tabhate}, IndicConan \cite{sahoo2024indicconan}, the Hindi--English code-mixed dataset \cite{bohra2018dataset}, the Bengali hate speech dataset \cite{romim2021hate}, and L3Cube-MahaHate \cite{velankar2022l3cube} are publicly available for research and benchmarking. However, licensing terms are often unspecified or inconsistently documented.

\paragraph{Topic Classification.}
Resources including IndicDialogue \cite{arnob2024indicdialogue}, TeClass \cite{kanumolu2024teclass}, L3Cube-IndicNews \cite{mirashi2023l3cube}, the multilingual factual-claim dataset \cite{dutta2022multilingual}, TamilMemes \cite{suryawanshi2020dataset}, MMT \cite{dalal2023mmt}, and SIB-200 \cite{adelani2024sib} are released as research datasets. Formal licensing information, however, is not uniformly specified.

\paragraph{Natural Language Understanding.}
NLU datasets such as BanglaParaphrase \cite{akil2022banglaparaphrase}, BnPC \cite{saha2024bnpc}, MahaParaphrase \cite{jadhav2025mahaparaphrase}, and the paraphrase corpus by \citet{singh2020creating}, along with benchmarks like IndicXNLI \cite{aggarwal2022indicxnli}, the code-mixed NLI dataset \cite{khanuja2020new}, and MILU \cite{verma2025milu}, are widely used for research and evaluation. Nonetheless, licensing terms are often undocumented or unclear.

\paragraph{Summarization.}
Summarization datasets such as MILDSum \cite{datta2023mildsum}, IndicSumm \cite{sireesha2023indicsumm}, PMIndiaSum \cite{urlana2023pmindiasum}, TeSum \cite{urlana2022tesum}, HindiSumm \cite{singh2024hindisumm}, L3Cube-MahaSum \cite{kulkarni2024l3cube}, Social-Sum-Mal \cite{rahul2024social}, COSMMIC \cite{kumar2025cosmmic} and Gupshup \cite{mehnaz2021gupshup} are typically released for academic research. However, licensing conditions vary and are not consistently documented across datasets.

\paragraph{Machine Translation.}
Machine translation resources, including PMIndia \cite{haddow2020pmindia}, IndicTrans2 \cite{gala2023indictrans2}, CorIL \cite{bhattacharjee2025coril}, EnIndic \cite{banerjee2023automatic}, and benchmarks such as IndicMT Eval \cite{dixit2023indicmt} and IL-ILGOV \cite{mujadia2025ilgov}, are generally distributed for research use. Licensing ranges from permissive to restricted, with some datasets lacking explicit terms.

\paragraph{Information Retrieval.}
Benchmarks such as Hindi-BEIR \cite{acharya2024hindi}, Anveshana \cite{jagadeeshan2025anveshana}, and CURE \cite{iqbal2021cure} are released for research and evaluation. Licensing information, however, is inconsistently specified across resources.

\paragraph{Dialogue Systems.}
Dialogue datasets including the code-mixed corpus \cite{banerjee2018dataset}, HDRS \cite{malviya2021hdrs}, TamilATIS \cite{ramaneswaran2022tamilatis}, and mTransDial \cite{ambastha2021mtransdial} are primarily available for academic use, though formal licensing documentation is often missing.

\paragraph{Speech Processing.}
Speech resources such as IndicSUPERB \cite{javed2023indicsuperb}, IndicVoices \cite{javed2024indicvoices}, IndicSpeech \cite{srivastava2020indicspeech}, and IndicST \cite{sethiya2025indic} are widely used in research. However, licensing varies across datasets and is not consistently reported.

\paragraph{Multimodal Language Understanding.}
Datasets such as Chitrakshara \cite{khan2025chitrakshara}, Bengali Visual Genome \cite{sen2022bengali}, M2H2 \cite{chauhan2021m2h2}, and INCLUDE \cite{sridhar2020include} are publicly released for research. Licensing terms remain heterogeneous and often undocumented.

\paragraph{Misinformation and Fact-Checking.}
Resources including FactDrill \cite{singhal2022factdrill}, FakeNewsIndia \cite{dhawan2022fakenewsindia}, FakeCovid \cite{shahi2020fakecovid}, TamilFacts \cite{francis2024tamilfacts}, and InDeepFake \cite{das2025indeepfake} are available for research use, though licensing terms are inconsistently specified.

\paragraph{Cultural NLP.}
Publicly available cultural NLP resources suitable for research include SANSKRITI \cite{maji2025sanskriti}, DRISHTIKON \cite{maji2025drishtikon}, Dosa \cite{seth2024dosa}, DIWALI \cite{sahoo2025diwali}, Pragyaan \cite{rachamalla2025pragyaan}, PARIKSHA \cite{watts2024pariksha}, and NativQA \cite{hasan2025nativqa}.

\paragraph{Emerging Topics.}
Publicly available resources for emerging areas such as bias and evaluation include IndiBias \cite{sahoo2024indibias}, Indian-BHED \cite{khandelwal2024indian}, ComMA \cite{kumar2022comma,kumar2024multilingual}, IndiCASA \cite{santhosh2025indicasa}, QUENCH \cite{khan2025quench}, ParamBench \cite{maheshwari2025parambench}, and IndicRAGSuite \cite{prasanjith2025indicragsuite}.

\section{Additional Survey Tables and Resources}

This appendix provides supplementary material that complements the main survey. In particular, we include detailed tabular summaries of prior work that could not be accommodated in the main text due to space constraints. These tables offer a fine-grained comparison of datasets, methodologies, and empirical findings across all seventeen tasks for Indian languages, serving as a consolidated reference for researchers.

\begin{figure*}[htbp]
    \centering
    \includegraphics[width=\linewidth]{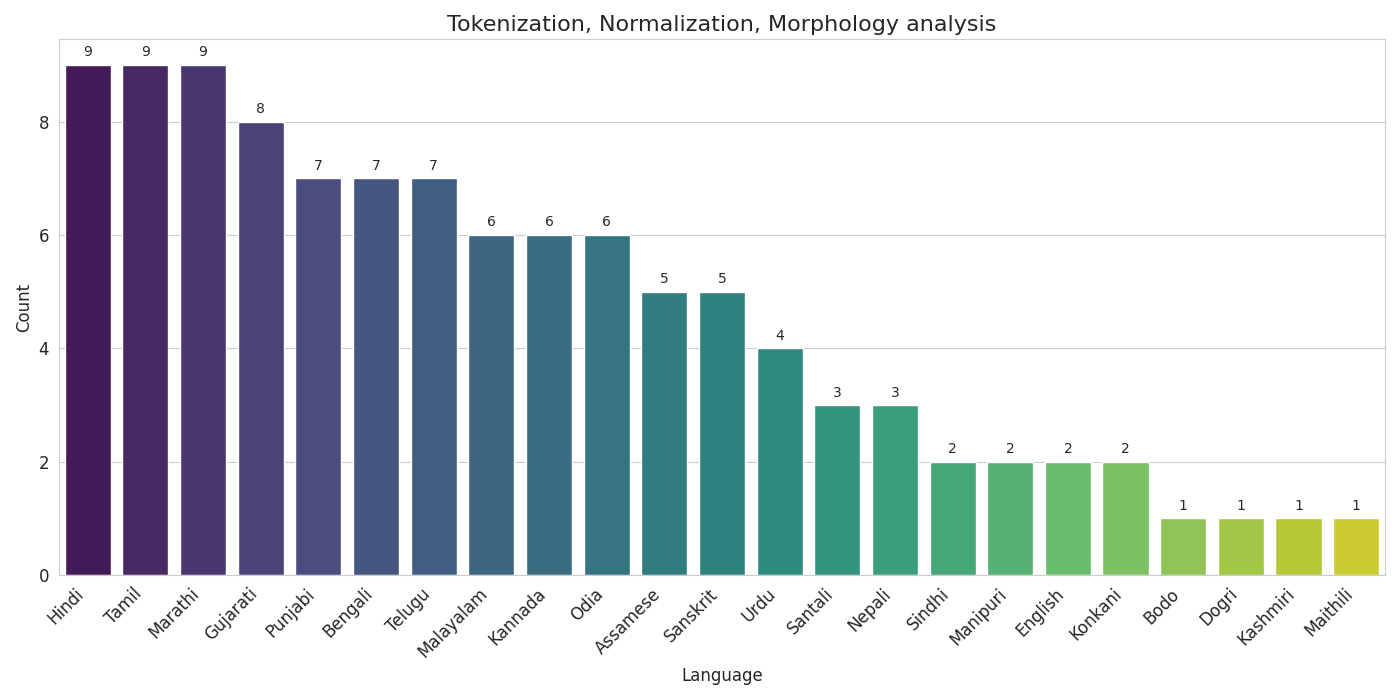}
    \caption{Language-wise distribution of datasets and studies focusing on tokenization, normalization, and morphological analysis across Indian languages.}
    \label{fig:language_count_tokenization}
\end{figure*}

\begin{table*}[htbp]
\centering
\small
\resizebox{\textwidth}{!}{
\begin{tabular}{p{2.6cm} p{3.2cm} p{2.8cm} p{4.2cm} p{4.5cm} p{4.2cm}}
\hline
\textbf{Paper} & \textbf{Focus \& Objective} & \textbf{Languages} & \textbf{Methodologies \& Algorithms} & \textbf{Key Resources / Contributions} & \textbf{Key Findings} \\
\hline

\cite{brahma2025morphtok} &
Morphology-aware tokenization for efficient LLM pre-training &
Hindi, Marathi &
Morphology-aware segmentation; Sandhi splitting; Constrained BPE (CBPE) &
New Hindi–Marathi dataset with sandhi annotations; EvalTok human evaluation metric &
CBPE reduced fertility by 1.68\%; improved MT and LM performance \\

\cite{ohm2024study} &
Tokenization analysis for a low-resource tribal language &
Santhali (Ol-Chiki) &
Character-, word-, and subword-level tokenization; spaCy-based models &
Empirical evaluation on Ol-Chiki paragraphs &
spaCy outperformed alternatives; highlighted script-specific properties \\

\cite{pattnayak2025tokenization} &
Impact of tokenization on zero-shot NER &
Assamese, Bengali, Marathi, Odia; Santali, Manipuri, Sindhi &
BPE vs. SentencePiece vs. Character tokenization; IndicBERT evaluation &
Intrinsic (OOV, morphology) vs. extrinsic (NER) analysis &
SentencePiece consistently outperformed BPE in zero-shot NER \\

\cite{kumar2024krutrim} &
Indic multilingual LLM data preparation and tokenizer design &
11 Indic languages &
Large-scale filtering and deduplication; multilingual tokenizer training &
Petabyte-scale corpus; custom Indic tokenizer for 3B/7B models &
Better token-to-word ratio than OpenAI Tiktoken \\

\cite{rana2025indicsupertokenizer} &
Efficient tokenizer for multilingual Indic LLMs &
22 Indic languages + English &
Subword + multiword tokenization; language-specific pre-tokenization &
IndicSuperTokenizer; extensive ablation studies &
39.5\% fertility improvement over LLaMA4; 44\% higher throughput \\

\cite{arora2020inltk} &
Indic NLP toolkit with tokenization and embeddings &
13 Indic languages + English &
Data augmentation; pretrained language models &
iNLTK library; pretrained models for 13 languages &
Achieved $>$95\% of SOTA using $<$10\% data \\

\cite{shahriar2024improving} &
Tokenization for morphologically rich languages &
Bengali, Hindi &
WordPiece vs. Unigram vs. Character BERT models &
Four pretrained BERT variants &
Unigram and character tokenizers outperformed WordPiece \\

\cite{saurav2020passage} &
Pretrained word embeddings for Indian languages &
14 Indic languages &
Contextual (BERT, ELMo) and non-contextual embeddings; cross-lingual training &
436 pretrained models across 8 approaches &
Contextual embeddings improved performance but were resource-intensive \\

\cite{akhtar2017word} &
Word similarity evaluation datasets &
Urdu, Telugu, Marathi, Punjabi, Tamil, Gujarati &
Manual translation and annotation; baseline evaluation &
Monolingual word similarity benchmarks &
Enabled standardized evaluation of word representations \\

\cite{gupta2022indic} &
Punctuation restoration and inverse text normalization &
11 Indic languages &
IndicBERT-based punctuation; WFST-based ITN &
Public indic-punct toolkit &
Improved readability and downstream ASR-based NLP tasks \\

\cite{krishnan2025normalized} &
Sanskrit dataset enrichment for morphology &
Sanskrit &
Integration of SH Segmenter and Saṃsādhanī tools &
Morphologically enriched DCS corpus &
Improved performance of Sanskrit segmenters \\

\cite{baxi2022gujmorph} &
Gujarati morphological dataset creation &
Gujarati &
Unimorph-based annotation; suffix analysis &
GujMORPH dataset (16,527 forms) &
First benchmark for Gujarati morphology \\

\cite{srirampur2014statistical} &
Statistical morphological analyzer &
Hindi, Urdu, Telugu, Tamil &
Feature-rich ML-based SMA++ &
Indic/Dravidian-specific feature engineering &
Outperformed Morfette and earlier SMA models \\

\cite{baxi2025bidirectional} &
Neural morphological analysis without rules &
Gujarati &
Bi-LSTM with alternate label representations &
Gold morphological dataset &
Improved accuracy without explicit suffix rules \\

\cite{premjith2018deep} &
Neural sandhi splitting and morphology &
Malayalam &
RNN, LSTM, GRU architectures &
Automatic morpheme segmentation systems &
GRU achieved best accuracy (98.16\%) \\

\cite{rajasekar2021comparison} &
Comparison of morphological analyzers &
Tamil (medical domain) &
Rule-based, paradigm-based, and n-gram analyzers &
Domain-specific evaluation corpus &
Paradigm-based Tacola achieved best results \\

\cite{singh2021morphological} &
Morphology-aware sentiment analysis &
Punjabi &
Morphological normalization + DNN classifier &
Farmer-suicide dataset &
Achieved 90.29\% sentiment accuracy \\

\cite{krishna2017dataset} &
Standardized Sanskrit word segmentation &
Sanskrit &
Formal objective definition; candidate generation &
115k sentence benchmark dataset &
Resolved inconsistencies in prior evaluations \\

\cite{sarveswaran2018thamizhifst} &
Tamil morphological analyzer and generator &
Tamil &
FST using FOMA; LFG-based modeling &
ThamizhiFST system &
Achieved F-measure of 0.97 for verbs \\

\cite{dasari2023transformer} &
Transformer-based morphology for low-resource language &
Telugu &
mBERT, XLM-R, IndicBERT vs. monolingual BERT-Te &
10k annotated sentences; monolingual BERT-Te &
Monolingual model outperformed multilingual ones \\

\cite{singh2016multiword} &
Multiword expression annotation &
Hindi, Marathi &
Corpus extraction + human annotation &
Gold MWE datasets &
Standard evaluation resource for MWEs \\

\cite{prathibha2013development} &
Hybrid morphological analyzer for MT &
Kannada &
Suffix stripping + rule-based + paradigm-based &
Evaluation on Kannada Rathna Kosha &
Effective MT-oriented morphological analysis \\

\hline
\end{tabular}
}
\caption{Summary of tokenization, normalization, and morphological analysis resources for Indian languages.}
\label{tab:tokenization_morphology}
\end{table*}

\begin{figure*}[htbp]
    \centering
    \includegraphics[width=\linewidth]{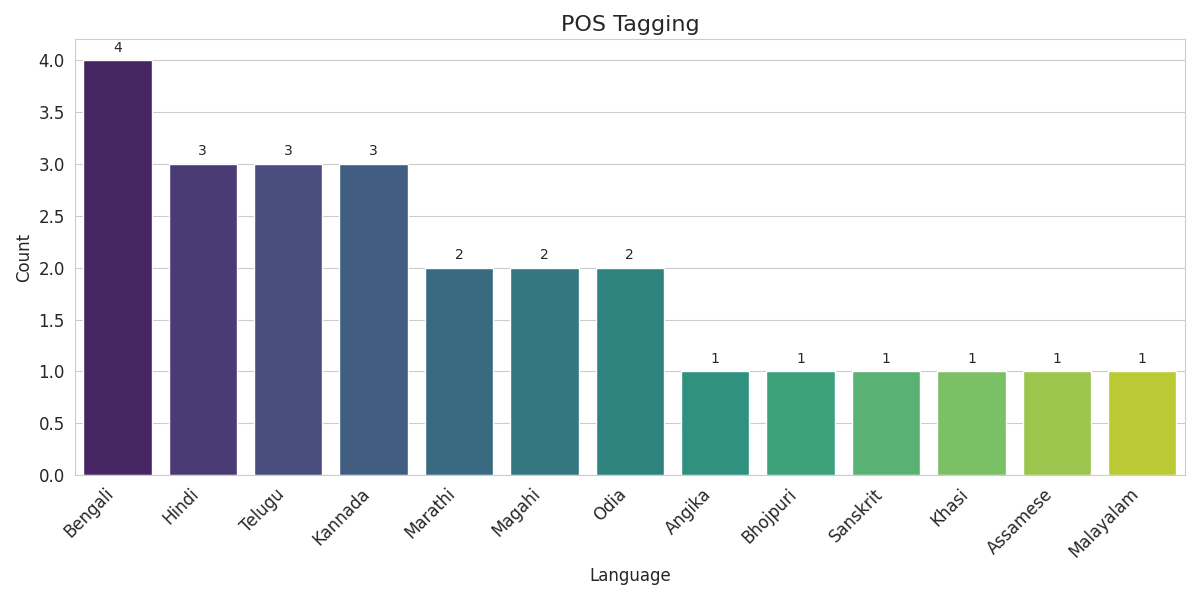}
    \caption{Language-wise distribution of datasets and studies focusing on POS-Tagging across Indian languages.}
    \label{fig:pos}
\end{figure*}

\begin{table*}[t]
\centering
\resizebox{\textwidth}{!}{
\begin{tabular}{p{3.5cm} p{4.5cm} p{5.5cm} p{5.5cm} p{5.5cm}}
\hline
\textbf{Focus \& Objective} & \textbf{Languages Covered} & \textbf{Methodologies \& Algorithms} & \textbf{Key Resources / Contributions} & \textbf{Key Findings} \\
\hline

Development of a trigram HMM-based POS tagger \cite{sarkar2013trigram} 
& Bengali, Hindi, Marathi, Telugu 
& Second-order Hidden Markov Model (HMM); prefix, suffix, and word-type analysis for unknown words 
& Implemented a trigram POS tagger from scratch 
& Portable across languages by replacing training data; performed comparably to or better than a bigram baseline \\

Deep learning POS tagging for a South Indian language \cite{rajani2022pos} 
& Kannada 
& Deep Neural Network combining word embeddings, RNN, and LSTM 
& TDIL dataset (10K annotated sentences / 190K words); BIS tagset (27 tags) 
& Achieved 81\% average accuracy on an unseen dataset \\

POS tagging for extremely low-resource languages \cite{kumar2024part} 
& Angika, Magahi, Bhojpuri, Hindi 
& Fine-tuning multilingual PLMs; zero-shot evaluation; proposed ``look-back fix'' for tokenization 
& First POS evaluation dataset for Angika; parallel dataset for four languages 
& Pretrained tokenizers underperform in zero-shot; look-back fix improved F1 by up to 8\% on Angika \\

Statistical and deep learning approaches for POS tagging \cite{dalai2023part} 
& Odia 
& CRF; CNN; Bi-LSTM with character sequence extraction 
& Mapped BIS tagset to Universal Dependencies (UD) tagset 
& Bi-LSTM with character features and pretrained embeddings achieved state-of-the-art results \\

Unsupervised POS tagging using deep learning \cite{srivastava2018deep} 
& Sanskrit 
& Vector-space representations; autoencoder for dimensionality reduction; Bi-LSTM autoencoder for clustering 
& Untagged Sanskrit corpus (JNU) for training; tagged corpus (115K words) for testing 
& Identified compressed representation dimensions yielding best clustering performance \\

CRF-based POS tagging for a low-resource language \cite{warjri2021part} 
& Khasi 
& Conditional Random Field (CRF) 
& Built Khasi corpus of $\sim$71K tokens; designed a dedicated tagset 
& Achieved 92.12\% accuracy and 0.91 F1-score on test data \\

Comparative study of sequential taggers \cite{kumar2012developing} 
& Magahi 
& SVM (SVMTool), HMM (TnT), Maximum Entropy (MxPost), Memory-Based (MBT) 
& Dataset of $\sim$50K training and $\sim$13K testing words using BIS tagset (33 tags) 
& Maximum Entropy tagger performed best after tuning; all models underperformed compared to English \\

Character-level deep learning models for POS tagging \cite{phukan2024exploring} 
& Assamese 
& Character-level LSTM and Bi-LSTM 
& Corpus of 60K words using LDCIL Assamese tagset 
& Character-level Bi-LSTM (93.36\%) outperformed LSTM (92.80\%) \\

POS tagging and chunking with pretrained transformers \cite{dalai2024deep} 
& Odia 
& RNN, CNN, Transformer-based models; word, character, and sub-word representations 
& Manually annotated Odia chunking dataset; custom chunking tagset 
& Transformer-based models achieved superior accuracy and robustness \\

Cross-language POS tagging using resource-rich languages \cite{reddy2011cross} 
& Kannada 
& Cross-lingual tagging; morphological analysis and lemmatization 
& Large Kannada corpora and morphological analyzer 
& Cross-language taggers matched monolingual accuracy and were faster than existing tools \\

Theoretical and technical analysis of POS tagging \cite{dash2013part} 
& Bengali 
& Rule-based schema design; hierarchical tag assignment modalities 
& Analysis based on Bengali written text corpus 
& Addressed theoretical challenges in lexico-semantic and grammatical function identification \\

Web-based corpus creation and POS tagging \cite{ekbal2008web} 
& Bengali, Hindi, Telugu 
& HMM and SVM 
& Bengali news corpus (34M wordforms); lexicon of 128K entries 
& SVM outperformed HMM across all languages (Bengali SVM accuracy: 91.23\%) \\

Unified parsing strategy with integrated POS tagging \cite{tandon2017unity} 
& Bengali, Marathi, Kannada, Telugu, Malayalam 
& Non-linear neural networks; monolingual distributed embeddings; suffix/postposition features 
& Built POS taggers and chunkers; first parsers for Marathi, Kannada, Malayalam 
& Embeddings captured morphological features (gender, number, person) without explicit analyzers \\

\hline
\end{tabular}
}
\caption{Summary of Part-of-Speech (POS) tagging approaches for Indian languages, covering statistical, neural, cross-lingual, and low-resource settings.}
\label{tab:pos_indian_languages}
\end{table*}

\begin{figure*}[htbp]
    \centering
    \includegraphics[width=\linewidth]{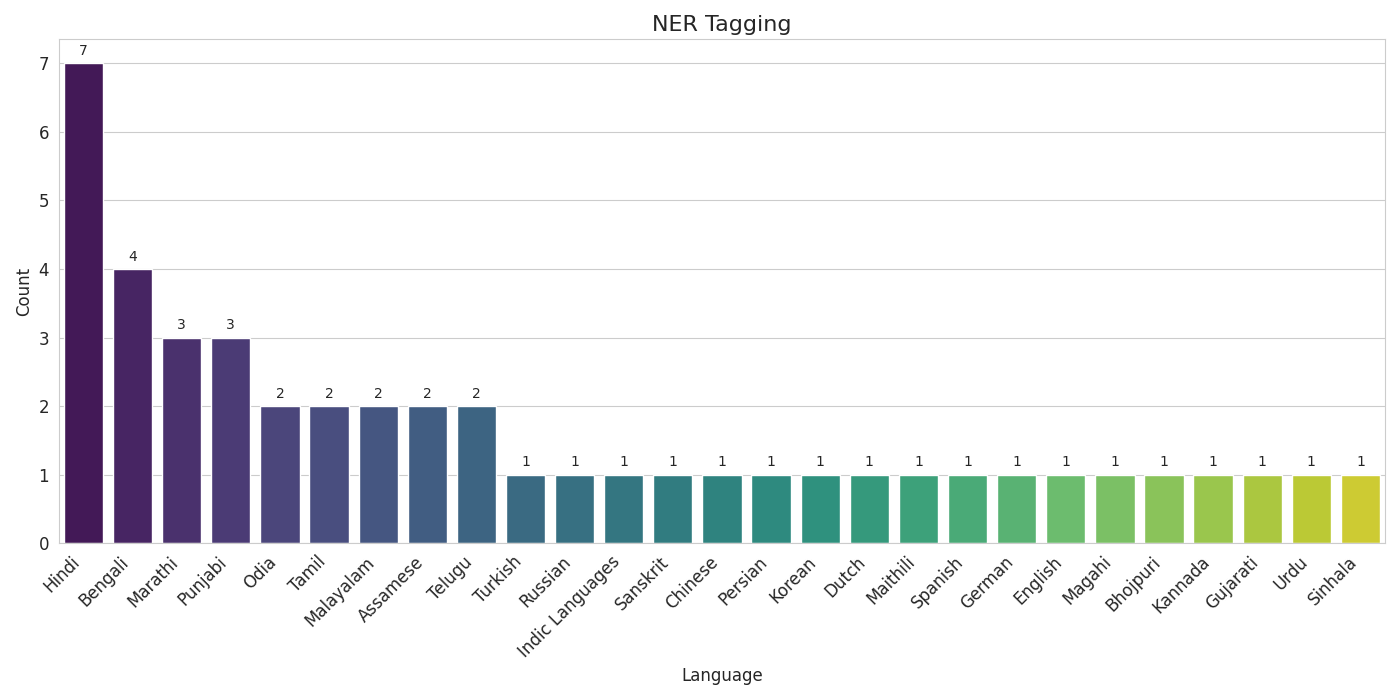}
    \caption{Language-wise distribution of datasets and studies focusing on Named Entity Recognition across Indian languages.}
    \label{fig:ner}
\end{figure*}

\begin{table*}[t]
\centering
\small
\setlength{\tabcolsep}{6pt}
\renewcommand{\arraystretch}{1.2}
\resizebox{\textwidth}{!}{
\begin{tabular}{p{4.8cm} p{3.2cm} p{4.8cm} p{4.8cm}}
\hline
\textbf{Work \& Focus} & \textbf{Languages} & \textbf{Methodologies} & \textbf{Key Resources / Contributions} \\
\hline

~\cite{bahad2024fine}: Transfer learning for adapting NER to Indian languages &
Hindi, Odia, Telugu, Urdu &
Fine-tuning pretrained transformers; cross-lingual transfer learning &
Annotated corpora of $\sim$40K sentences across 4 languages; multilingual fine-tuned NER model \\

~\cite{murthy2022hiner}: Large-scale standard-compliant Hindi NER dataset creation &
Hindi &
Sequence labeling with multiple language models for benchmarking &
\textbf{HiNER}: 109,146 sentences, 2.2M tokens, 11 entity tags \\

~\cite{mhaske2023naamapadam}: Multilingual Indic NER dataset via projection &
11 Indic languages &
Projection from English using Samanantar; IndicBERT fine-tuning &
\textbf{Naamapadam}: $>$400K sentences, $>$100K entities; IndicNER model \\

~\cite{sharma2022named}: MuRIL-based Hindi NER modeling &
Hindi &
MuRIL representations with CRF; layer-wise fine-tuning analysis &
Hindi NER system trained on ICON 2013 dataset \\

~\cite{mundotiya2023development}: NER for low-resource Purvanchal languages &
Bhojpuri, Maithili, Magahi &
LSTM--CNN--CRF; CRF baselines &
Annotated corpora (228K/157K/56K tokens); 22 entity tags \\

~\cite{malmasi2022multiconer}: Complex multilingual NER in low-context settings &
Multilingual (incl. Hindi, Bengali) &
XLM-RoBERTa; gazetteer-augmented GEMNET &
\textbf{AnonData}: 26M-token dataset across Wiki, queries, and short texts \\

~\cite{sujoy2023pre}: Reducing annotation cost for Sanskrit NER &
Sanskrit &
Heuristic pre-annotation using transliteration and gazetteers &
Pre-annotated corpus from \textit{Srimad-Bhagavatam} \\

~\cite{sharma2020deep}: Efficient deep neural architecture for Hindi NER &
Hindi &
CNN + BiLSTM + CRF; word and character embeddings &
Lightweight DNN architecture for resource-scarce Hindi \\

~\cite{goyal2021deep}: Embedding-enhanced NER without heavy feature engineering &
Hindi, Punjabi &
Bi-GRU + CNN; Enhanced Word Embeddings (FastText + linguistic cues) &
Punjabi NER dataset; Hindi evaluation on IJCNLP-08 NERSSEAL \\

~\cite{mohan2023fine}: Multilingual BERT fine-tuning for Indian NER &
Tamil, Malayalam, Bengali, Marathi &
Multilingual BERT fine-tuning &
NER benchmarks built on WikiAnn \\

~\cite{litake2022l3cube}: Gold-standard Marathi NER benchmark &
Marathi &
CNN, LSTM, mBERT, XLM-R, IndicBERT, MahaBERT &
\textbf{L3Cube-MahaNER}: First large-scale Marathi gold dataset \\

~\cite{haque2023b}: Balanced Bangla NER with diverse entity types &
Bangla &
BiLSTM + fastText; sentence transformers; IndicBERT &
\textbf{B-NER}: 22,144 sentences, 8 entity types \\

~\cite{pathak2022asner}: Assamese NER dataset and baselines &
Assamese &
BiLSTM-CRF; FastText, BERT, XLM-R, MuRIL, FLAIR &
\textbf{AsNER}: $\sim$99K tokens from speeches and plays \\

~\cite{dahanayaka2014named}: Early data-driven Sinhala NER study &
Sinhala &
CRF; Maximum Entropy &
First NER resources and baselines for Sinhala \\

~\cite{singh2023deepspacy}: Deep learning-based Punjabi NER &
Punjabi (Gurmukhi) &
spaCy-based neural NER; Doccano annotation &
15K-sentence annotated benchmark corpus \\

\hline
\end{tabular}
}
\caption{Representative Named Entity Recognition (NER) resources and modeling efforts across Indian and neighboring languages, covering dataset creation, low-resource settings, and multilingual transformer-based approaches.}
\label{tab:indic_ner_resources}
\end{table*}

\begin{figure*}[htbp]
    \centering
    \includegraphics[width=\linewidth]{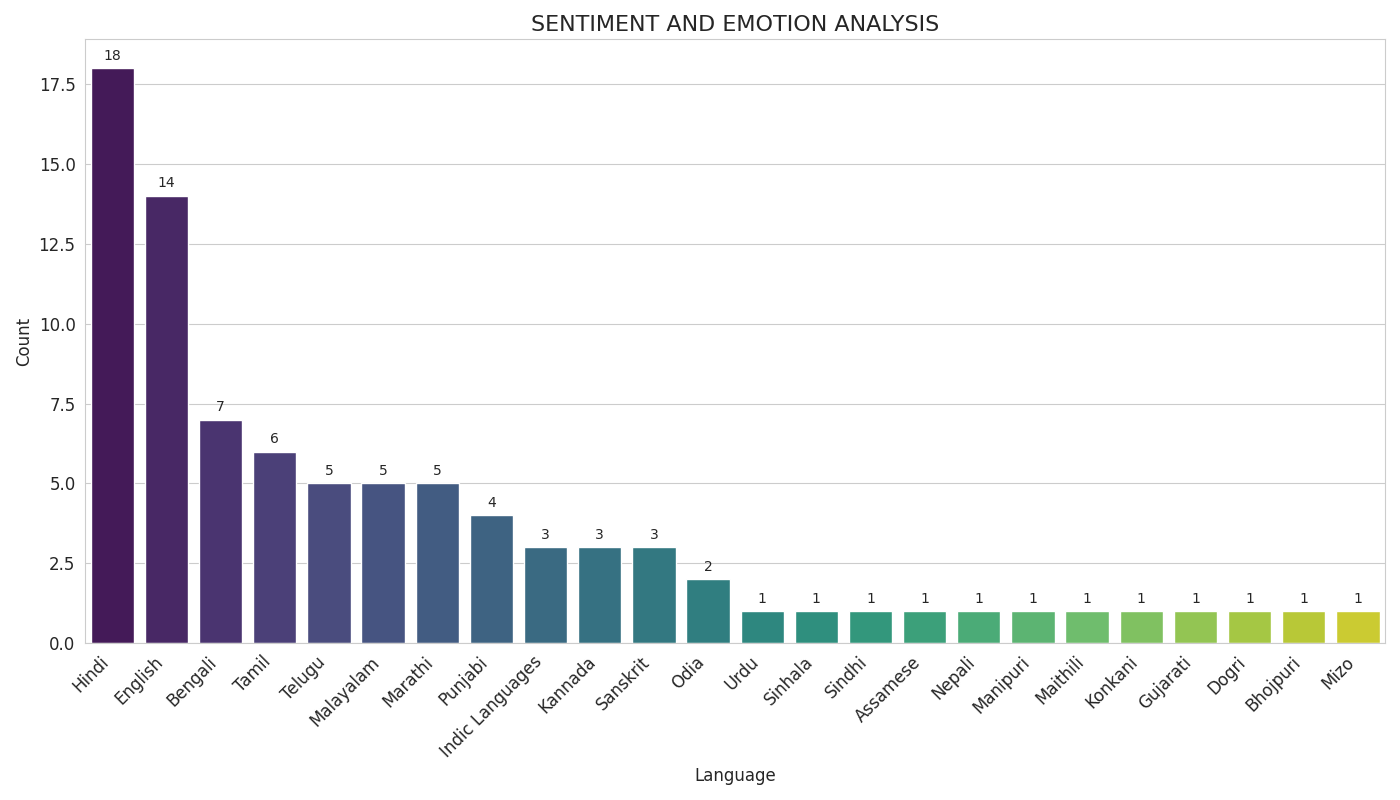}
    \caption{Language-wise distribution of datasets and studies focusing on Sentiment and Emotion Analysis across Indian languages.}
    \label{fig:senti}
\end{figure*}

\begin{table*}[t]
\centering
\resizebox{\textwidth}{!}{
\begin{tabular}{p{3cm} p{4.2cm} p{4cm} p{4.5cm} p{5.2cm}}
\hline
\textbf{Paper} & \textbf{Focus \& Objective} & \textbf{Languages Covered} & \textbf{Methodologies \& Algorithms} & \textbf{Key Resources / Contributions} \\
\hline

\cite{kumar2024indisentiment140} &
Sentiment dataset creation for low-resource languages via MT &
22 Indian languages &
Machine Translation (Google Translate) of Sentiment140 &
IndiSentiment140 multilingual sentiment dataset \\

\cite{chakravarthi2020sentiment} &
Sentiment analysis for code-mixed social media &
Malayalam--English &
Manual annotation; supervised vs.\ unsupervised comparison &
Gold-standard Malayalam--English code-mixed corpus \\

\cite{phani2016sentiment} &
Tweet sentiment analysis with simple features &
Bengali, Hindi, Tamil &
Language-independent features &
Used SAIL shared-task dataset \\

\cite{patra2015shared} &
Overview of Twitter sentiment shared task &
Bengali, Hindi, Tamil &
Constrained and unconstrained shared-task systems &
SAIL Tweets Dataset \\

\cite{kumar2021sentiment} &
Zero-shot transfer learning for sentiment analysis &
Hindi (from English) &
XLM-RoBERTa; zero-shot transfer &
IITP-Movie and IITP-Product datasets \\

\cite{chakravarthi2022dravidiancodemix} &
Sentiment and offensive language detection &
Tamil-Eng, Kannada-Eng, Malayalam-Eng &
ML and DL baselines &
DravidianCodeMix: $\sim$60K YouTube comments \\

\cite{kannadaguli2021code} &
Sentiment analysis for code-diverse text &
Kannada--English &
Supervised ML/DL models &
First Kannada--English code-mixed corpus \\
\hline
\end{tabular}
}
\caption{Summary of sentiment and emotion analysis resources for Indian languages.}
\label{tab:sentiment_emotion_indic_appendix}
\end{table*}

\begin{figure*}[htbp]
    \centering
    \includegraphics[width=\linewidth]{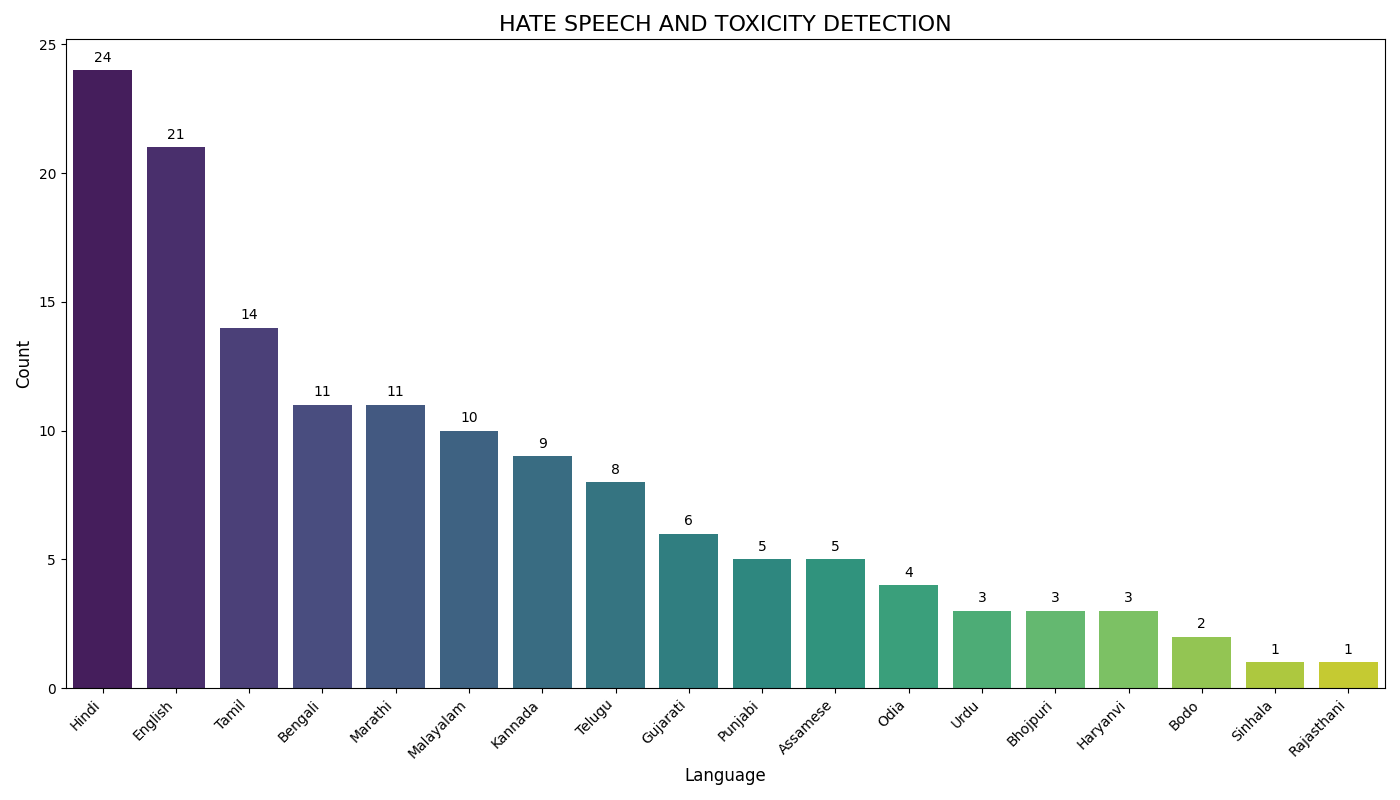}
    \caption{Language-wise distribution of datasets and studies focusing on Hate Speech and Toxicity Detection across Indian languages.}
    \label{fig:hate}
\end{figure*}

\begin{table*}[t]
\centering
\resizebox{\textwidth}{!}{
\begin{tabular}{p{3.2cm} p{4.6cm} p{3.6cm} p{5.2cm} p{5.8cm}}
\hline
\textbf{Paper (Citation)} & \textbf{Focus \& Objective} & \textbf{Languages Covered} & \textbf{Methodologies \& Algorithms} & \textbf{Key Resources / Contributions} \\
\hline

\cite{chakravarthi2020hopeedi} &
Hope speech detection emphasizing encouraging and inclusive content &
English, Tamil, Malayalam &
Manual annotation; Inter-annotator agreement (Krippendorff’s $\alpha$); Baseline classification &
HopeEDI dataset with 59K+ annotated YouTube comments \\

\cite{gupta2022adima} &
Audio-based abusive content detection to avoid ASR latency issues &
Hindi, Bengali, Punjabi, Haryanvi, Kannada, Odia, Bhojpuri, Gujarati, Tamil, Marathi &
Monolingual and cross-lingual zero-shot learning on speech data &
ADIMA dataset: 11,775 audio samples (65 hours) \\

\cite{sharma2024tabhate} &
Target-based hate speech detection &
Hindi &
Transformer-based deep learning models &
TABHATE: Hindi dataset annotated for targets (individual/group/community) \\

\cite{bohra2018dataset} &
Hate speech detection in code-mixed tweets &
Hindi--English (Code-mixed) &
Supervised models using character-, word-, and lexicon-level features &
Annotated Twitter dataset with word-level language tags \\

\cite{sahoo2024indicconan} &
Counter-narrative generation against hate speech &
Hindi, Indian English &
Human-in-the-loop auto-regressive generation with correction cycles &
IndicCONAN dataset with 2.5K+ CN examples per language \\

\cite{biradar2021hate} &
Translation/transliteration-based hate speech identification &
Hinglish &
IndicNLP transliteration; mBERT features; TIF-DNN &
Transformer-based interpretation framework for code-mixed data \\

\cite{gupta2025caste} &
Caste-based hate speech detection &
Hindi &
Hybrid expert + LLM annotation; MuRIL benchmarking &
First curated Hindi caste-hate dataset \\
\hline
\end{tabular}
}
\caption{Overview of Hate Speech and Toxicity Detection Resources for Indian Languages}
\label{tab:hate_speech_indic}
\end{table*}

\begin{figure*}[htbp]
    \centering
    \includegraphics[width=\linewidth]{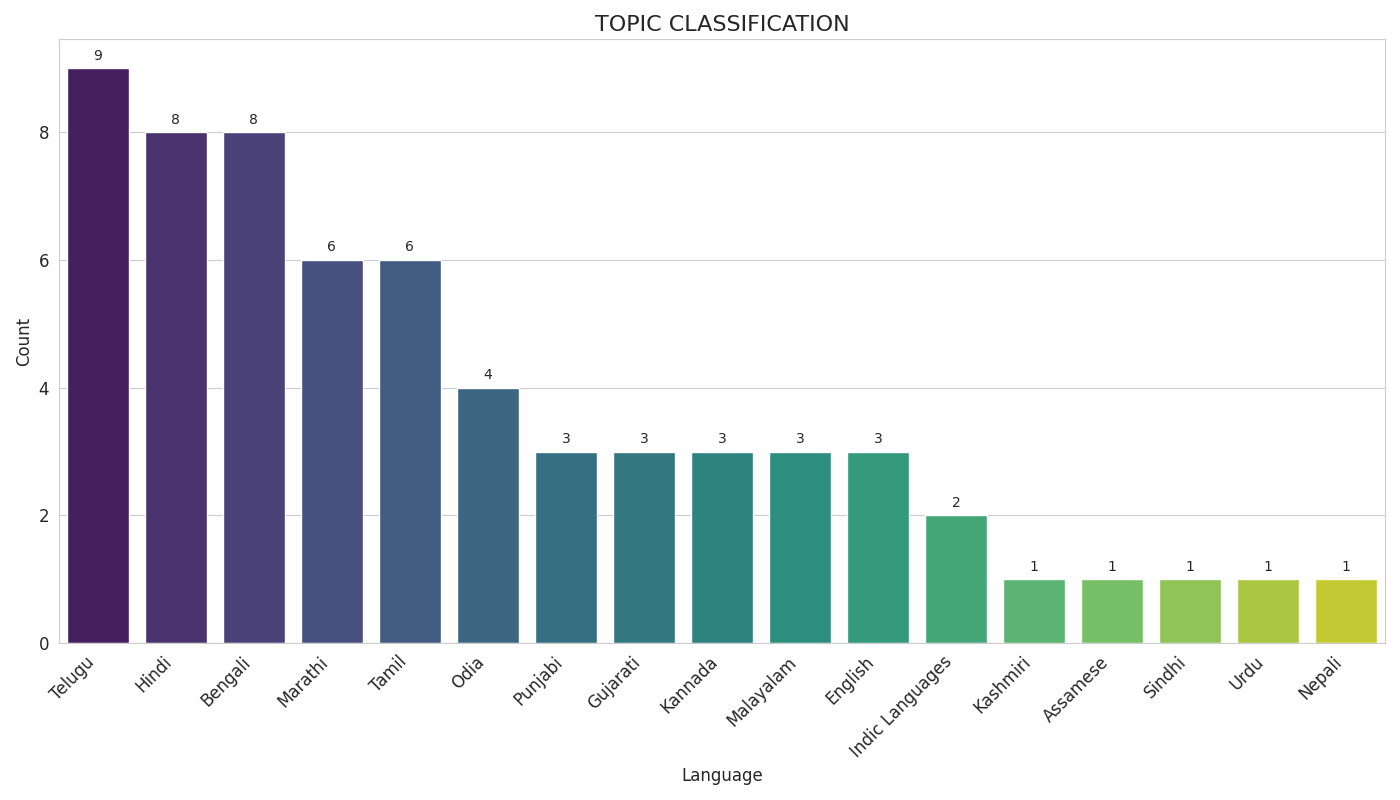}
    \caption{Language-wise distribution of datasets and studies focusing on Topic Classification across Indian languages.}
    \label{fig:topic}
\end{figure*}

\begin{table*}[t]
\centering
\resizebox{\textwidth}{!}{
\begin{tabular}{p{3cm} p{3.6cm} p{3.4cm} p{4.2cm} p{4.2cm} p{4.2cm}}
\hline
\textbf{Paper} & \textbf{Focus \& Objective} & \textbf{Languages Covered} & \textbf{Methodologies \& Algorithms} & \textbf{Key Resources / Contributions} & \textbf{Key Findings} \\
\hline

\cite{mirashi2023l3cube} &
Dataset creation for language modeling and downstream tasks such as topic modeling and conversation synthesis. &
Hindi, Bengali, Marathi, Telugu, Tamil, Urdu, Odia, Sindhi, Nepali, Assamese. &
Data sourced from OpenSubtitles.org; preprocessing to remove tags and timestamps; JSONL storage. &
IndicDialogue dataset with 7,750 SRT files, $\sim$6.85M dialogues, and $\sim$42.18M words. &
Addresses low representation of Indic languages and supports pre-training and downstream tasks. \\

\cite{deyar2025dataset} &
Benchmarking news snippet classification in a low-resource setting. &
Kashmiri. &
English-to-Kashmiri translation; manual refinement; fine-tuning ParsBERT and LLMs. &
15,036 manually labeled Kashmiri news snippets across 10 domains. &
ParsBERT achieves F1 = 0.98; first manually labeled Kashmiri news corpus. \\

\cite{dutta2022multilingual} &
Fine-grained identification of factual claims in social media. &
English, Bengali, Hindi, code-mixed. &
Token-level annotation of tweets. &
Multilingual fact-checking dataset for Indian Twitter. &
Highlights challenges of multilingual and code-mixed fact-checking. \\

\cite{suryawanshi2020dataset} &
Classification of memes as troll or non-troll. &
Tamil. &
Image-based meme classification. &
TamilMemes annotated dataset. &
Image-only approaches are insufficient for troll detection. \\

\cite{kunchukuttan2020ai4bharat} &
Creation of large monolingual corpora and word embeddings. &
10 Indic languages including Hindi, Bengali, Tamil, Telugu. &
Training word embeddings; evaluation via news classification. &
IndicNLP corpus with 2.7B words and pretrained embeddings. &
Indic embeddings outperform publicly available alternatives. \\
\hline
\end{tabular}
}
\caption{Representative topic classification and document categorization resources for Indian settings.}
\label{tab:topic_classification_indic}
\end{table*}

\begin{figure*}[htbp]
    \centering
    \includegraphics[width=\linewidth]{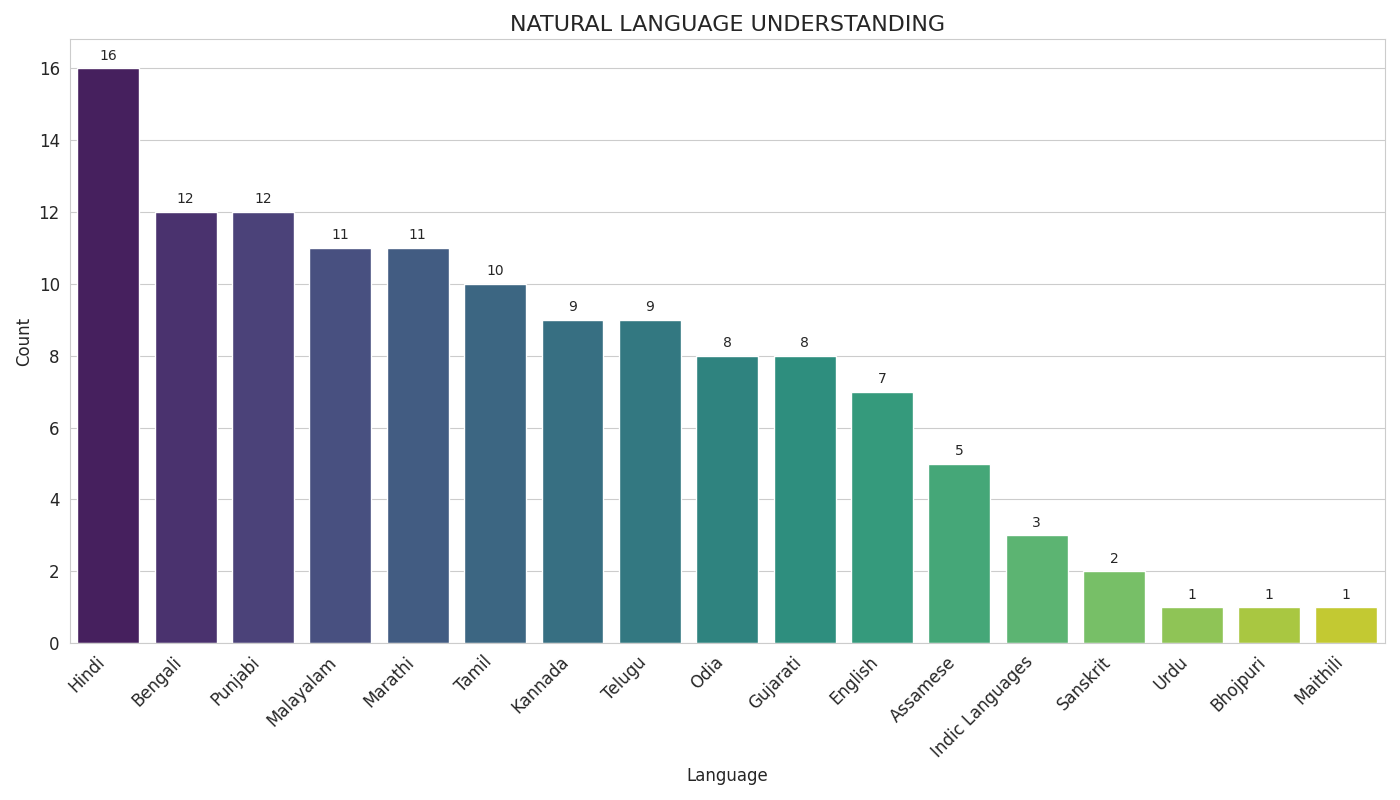}
    \caption{Language-wise distribution of datasets and studies focusing on Natural Language Understanding across Indian languages.}
    \label{fig:nlu}
\end{figure*}

\begin{table*}[t]
\centering
\resizebox{\textwidth}{!}{
\begin{tabular}{p{3.2cm} p{3.4cm} p{3.8cm} p{5.2cm} p{5.4cm}}
\hline
\textbf{Paper} & \textbf{Focus \& Objective} & \textbf{Languages Covered} & \textbf{Methodologies \& Algorithms} & \textbf{Key Resources / Contributions} \\
\hline
\cite{bhargava2017deep} & Paraphrase Detection & Hindi, English, Malayalam, Tamil, Punjabi & CNN with word embeddings, CNN+WordNet, RNN (LSTM, Bi-LSTM) & Early deep learning approaches for paraphrase identification \\

\cite{singh2020creating} & Paraphrase Corpus Creation & Hindi, Tamil, Malayalam, Punjabi & Manual annotation with two-stage linguistic verification & DPIL: First public paraphrase corpus for these languages \\

\cite{jadhav2025mahaparaphrase} & Paraphrase Detection & Marathi & Transformer-based BERT models & L3Cube-MahaParaphrase: 8K annotated sentence pairs \\

\cite{sethi2016novel} & Paraphrase Generation & Hindi (title), English (abstract) & Rule-based syntactic transformations, synonym/antonym replacement & Algorithm for structure-preserving paraphrase generation \\

\cite{anagha2023paraphrase} & Paraphrase Generation \& Detection & Kannada & Deep learning-based framework & Simplification-oriented paraphrase framework for low-resource settings \\

\cite{gupta2025knparaphraser} & Paraphrase Generation & Kannada & RTT, Seq2Seq NMT, T5-based model & KnParaphraser model and KanPara (100K sentence corpus) \\

\cite{siripragada2020multilingual} & Parallel Corpora Collection & 10 Indic + English & Web mining, DNN-based sentence alignment & Large multilingual sentence-aligned corpora \\
\hline
\end{tabular}
}
\caption{Representative datasets, benchmarks, and models for paraphrasing, inference, and semantic similarity in Indian languages.}
\label{tab:nlu-indic-resources}
\end{table*}

\begin{figure*}[htbp]
    \centering
    \includegraphics[width=\linewidth]{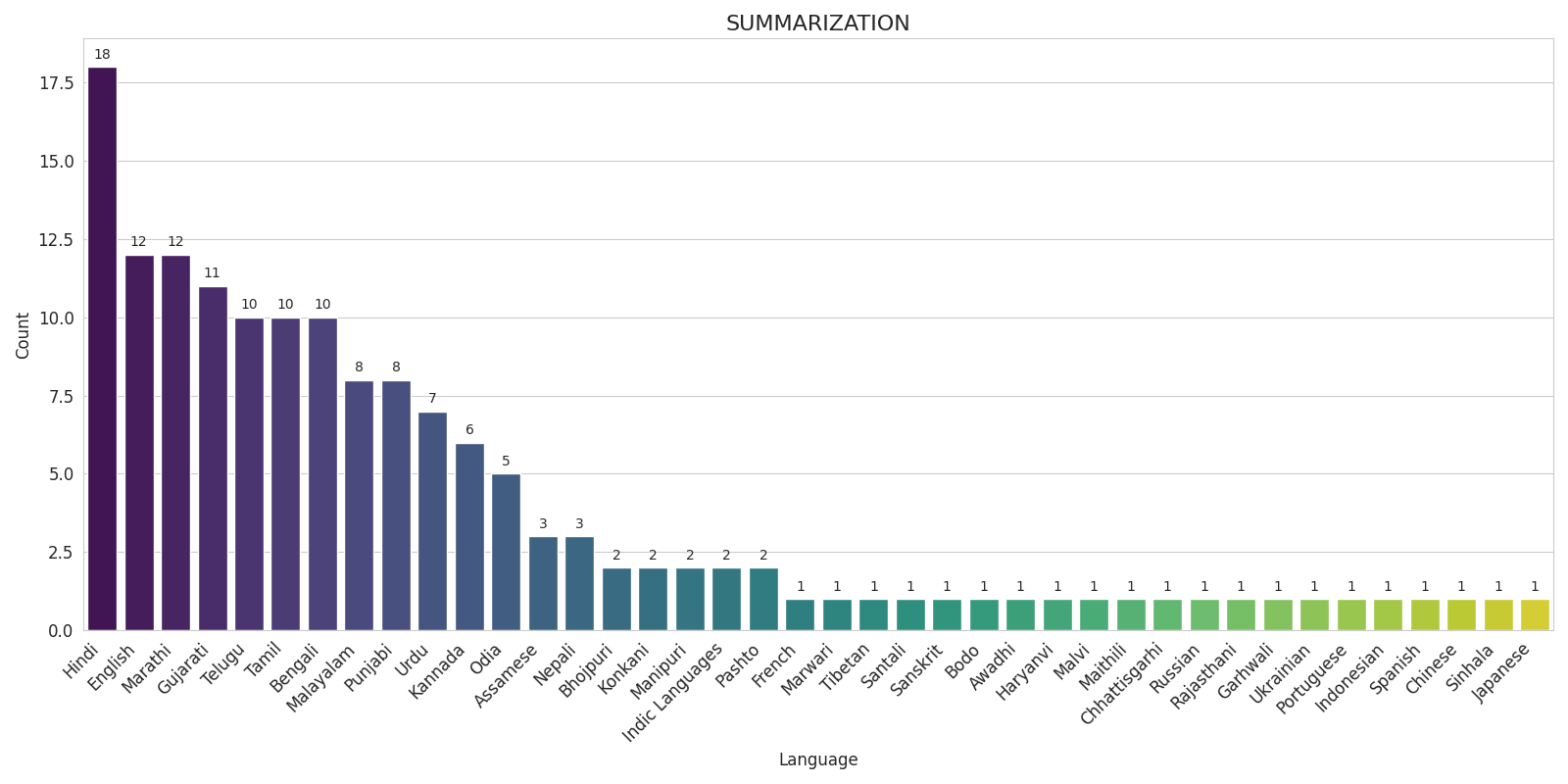}
    \caption{Language-wise distribution of datasets and studies focusing on Summarization across Indian languages.}
    \label{fig:summ}
\end{figure*}

\begin{table*}[t]
\centering
\resizebox{\textwidth}{!}{
\begin{tabular}{p{3.2cm} p{3.6cm} p{3.2cm} p{4.2cm} p{4.6cm}}
\hline
\textbf{Paper} & \textbf{Focus \& Objective} & \textbf{Languages Covered} & \textbf{Methodologies \& Algorithms} & \textbf{Key Resources / Contributions} \\
\hline
\cite{datta2023mildsum} &
Cross-lingual summarization of legal case judgments &
English, Hindi &
Benchmarked diverse summarization approaches &
MILDSum dataset with 3,122 legal case judgments and bilingual summaries \\

\cite{kulkarni2024l3cube} &
Abstractive news summarization &
Marathi &
IndicBART fine-tuning &
MahaSum dataset with 25K verified news summaries \\

\cite{khan2025automatic} &
Review of word embeddings in ATS &
Hindi &
Analysis of conventional, contextual, and multilingual embeddings &
Critical survey highlighting dataset scarcity \\

\cite{jain2021text} &
Extractive summarization &
Punjabi &
Three-phase neural network architecture &
Punjabi corpus derived from ILCI Phase-II \\

\cite{kumar2025cosmmic} &
Comment-sensitive multimodal summarization &
Hindi, Bengali, Marathi, Gujarati, Malayalam, Odia, Tamil, Telugu, Kannada &
LLaMA3, GPT-4, IndicBERT, CLIP &
COSMMIC dataset with articles, images, and reader comments \\

\cite{pimpalshende2024multilingual} &
Text and audio summarization for education &
Indian languages &
Script- and grammar-aware NLP pipelines &
End-to-end system for text and voice summarization \\

\cite{munaf2024low} &
Low-resource abstractive summarization &
Urdu &
Self-attentive transformers (mBERT, mT5 → urT5) &
76.5K Urdu article–summary pairs \\
\hline
\end{tabular}
}
\caption{Summary of datasets, benchmarks, and methods for text, multimodal, and multilingual summarization in Indian languages.}
\label{tab:indic_summarization_resources}
\end{table*}

\begin{figure*}[htbp]
    \centering
    \includegraphics[width=\linewidth]{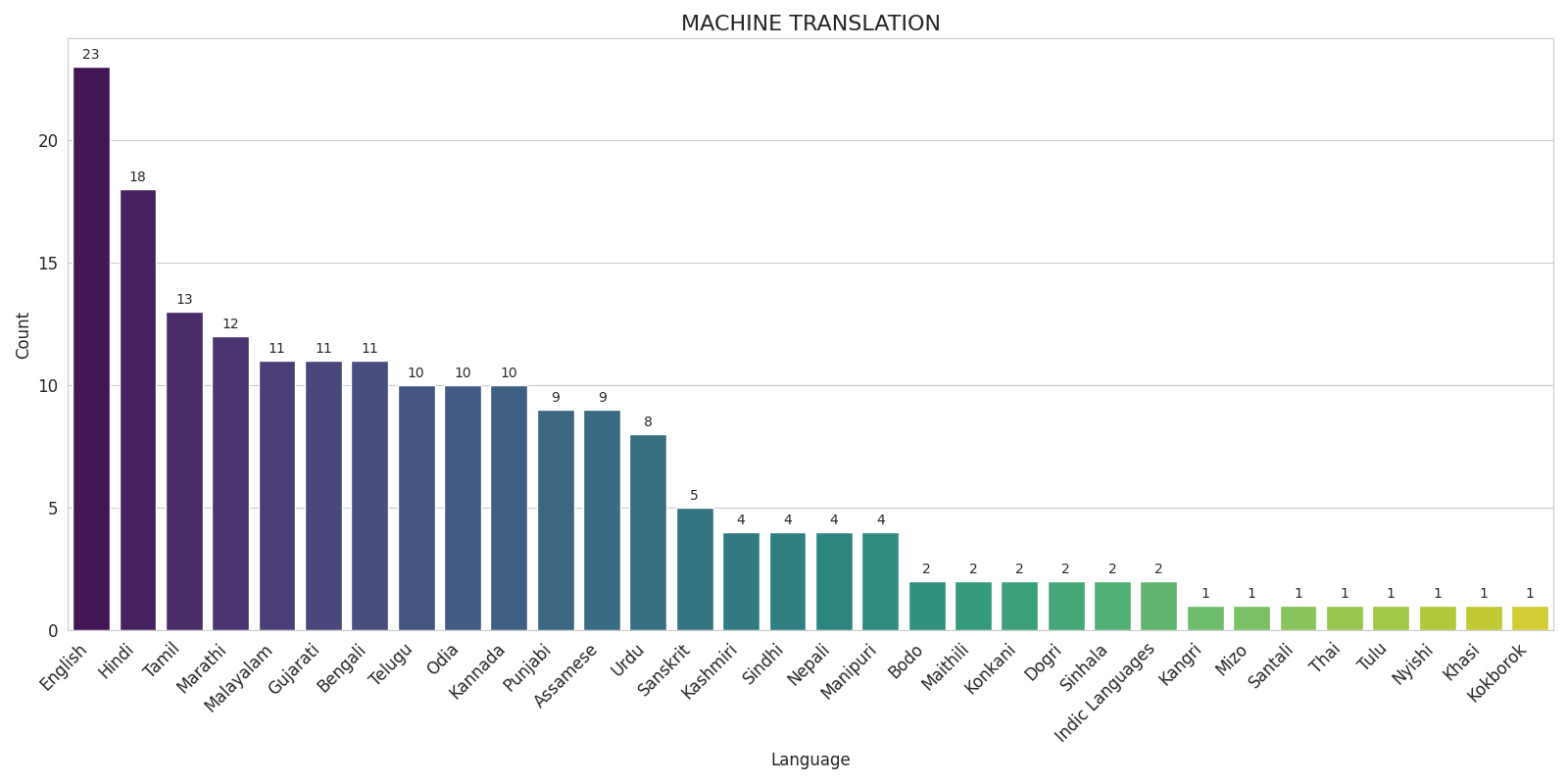}
    \caption{Language-wise distribution of datasets and studies focusing on Machine Translation across Indian languages.}
    \label{fig:mt}
\end{figure*}

\begin{table*}[t]
\centering
\resizebox{\textwidth}{!}{
\begin{tabular}{p{2.8cm} p{4.2cm} p{4.2cm} p{5.2cm} p{5.2cm}}
\hline
\textbf{Paper} & \textbf{Focus \& Objective} & \textbf{Languages Covered} & \textbf{Methodologies \& Algorithms} & \textbf{Key Resources / Contributions} \\
\hline
\cite{baruah2021low} & Comparative experiments on Assamese MT (low-resource). & Assamese, Bangla, Gujarati, Hindi, Marathi, Odia, Sinhalese, Urdu & Phrase-based SMT; NMT (Seq2Seq + Attention, Transformer, fine-tuned Transformer) & First baseline MT work involving Assamese; comparative evaluation \\
\cite{choudhary2020neural} & NMT for morphologically rich languages to overcome OOV issues. & English--Tamil, English--Malayalam & Multi-head self-attention; pre-trained BPE and MultiBPE embeddings & Curated and refined parallel corpora \\
\cite{sethi2023novel} & Feasibility of Sanskrit--Hindi MT (ancient to modern). & Sanskrit, Hindi & SMT (Moses); CNN-based NMT; encoder--decoder with attention + GRU & Novel DL model; manually created parallel corpus \\
\cite{bhattacharjee2025coril} & Enriching parallel corpora and analyzing domain sensitivity. & English, Telugu, Hindi, Punjabi, Odia, Kashmiri, Sindhi, Dogri, Kannada, Urdu, Gujarati & Fine-tuning IndicTrans2, NLLB, BhashaVerse & CorIL: 772K bi-text pairs across government, health, and general domains \\
\cite{jain2020neural} & Automated Tamil--English MT system. & Tamil, English & Encoder--decoder architectures; pre-trained embeddings; hyperparameter tuning & Released high-quality benchmark corpus \\
\cite{jain2024bhasaanuvaad} & Automatic Speech Translation (AST) for Indian languages. & 13 Indic languages + English & AST benchmarking; web mining; synthetic data generation & Bhasaanuvaad: 44.4K hours, 17M speech segments \\
\cite{lalrempuii2023extremely} & Extremely low-resource MT using multilingual techniques. & Mizo, English (+13 Indic languages) & Multilingual NMT; ensemble decoding; transliteration & Multilingual framework for Mizo MT \\
\hline
\end{tabular}
}
\caption{Representative Machine Translation datasets, models, and benchmarks for Indian languages.}
\label{tab:indic_mt}
\end{table*}

\begin{figure*}[htbp]
    \centering
    \includegraphics[width=\linewidth]{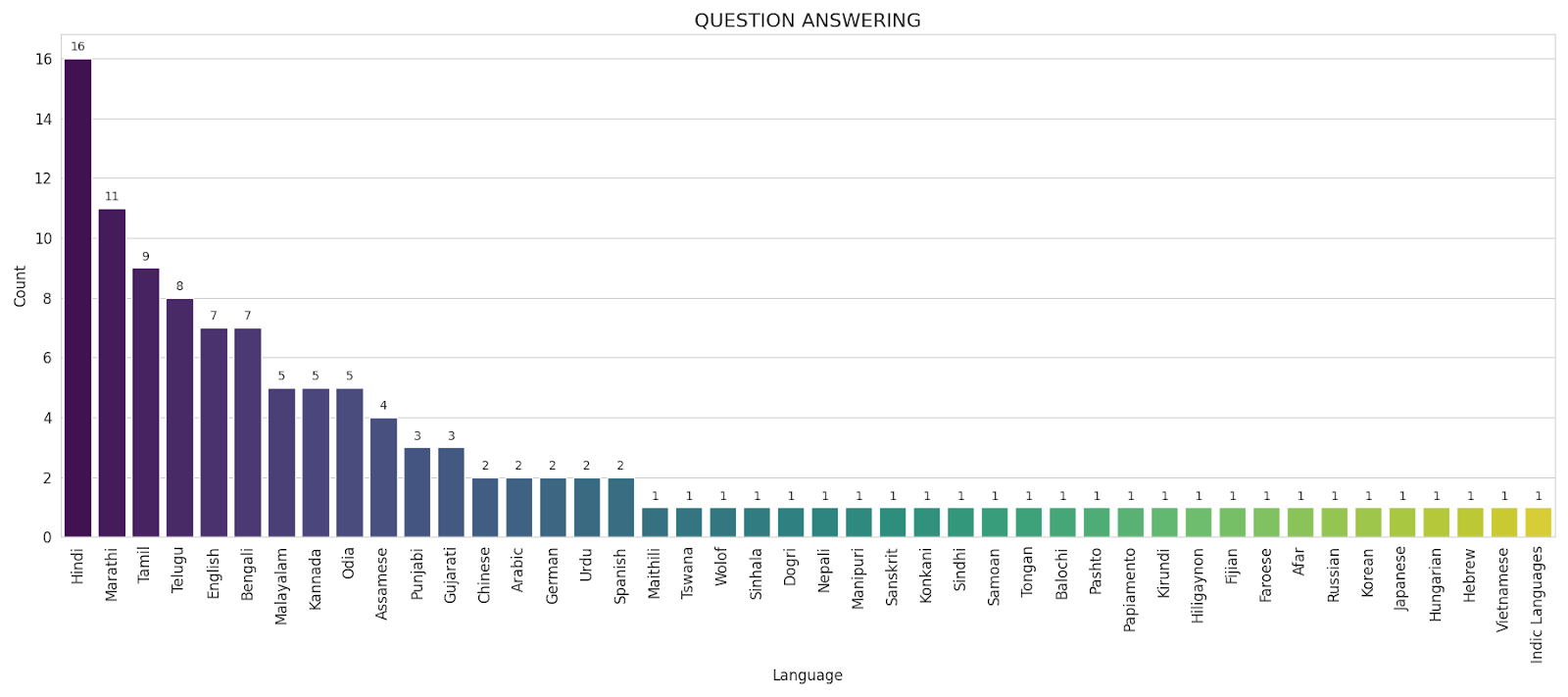}
    \caption{Language-wise distribution of datasets and studies focusing on Question Answering across Indian languages.}
    \label{fig:qa}
\end{figure*}

\begin{table*}[t]
\centering
\resizebox{\textwidth}{!}{
\begin{tabular}{p{3.2cm} p{4.2cm} p{3.6cm} p{5.6cm} p{5.8cm}}
\hline
\textbf{Paper} & \textbf{Focus \& Objective} & \textbf{Languages Covered} & \textbf{Methodologies \& Algorithms} & \textbf{Key Resources / Contributions} \\
\hline

\cite{rahmath2025transqam} & Extractive QA for exact answer span extraction & Malayalam & BERT, mBERT, XLM-RoBERTa, MuRIL & TransQAM dataset (30K QA pairs); SQuAD-style Malayalam QA system \\

\cite{ajawan2024krishiq} & Few-shot closed-domain agricultural QA & Kannada & Retriever–Reader pipeline (Haystack + DistilBERT) & Krishiq agricultural dataset; Krishiq-BERT model \\

\cite{lewis2020mlqa} & Cross-lingual extractive QA evaluation & EN, ES, DE, AR, HI, VI, ZH & Cross-lingual QA benchmarking; MT-based baselines & MLQA benchmark with multi-way aligned QA instances \\

\cite{tomar2025bharatbbq} & Social bias evaluation in LMs & 8 Indic + English & Zero-/few-shot evaluation; translation and verification & BharatBBQ benchmark with 392K bias instances \\

\cite{gupta2018mmqa} & Multi-domain multilingual QA framework & Hindi, English & DL-based question classification; similarity-based ranking & MMQA corpus (5,495 QA pairs) and framework \\

\cite{kumar2022mucot} & QA for low-resource languages via contrastive learning & 7 Indic + English & mBERT fine-tuning with contrastive loss & Translation-based augmentation; public code release \\

\cite{vats2025multilingual} & Structured QA with long-context modeling & Hindi, Marathi & State Space Models (SSMs) & First application of SSMs to Indic QA \\
\hline
\end{tabular}
}
\caption{Overview of Question Answering (QA) datasets, systems, and benchmarks for Indic and multilingual settings.}
\label{tab:indic_qa_overview}
\end{table*}

\begin{figure*}[htbp]
    \centering
    \includegraphics[width=\linewidth]{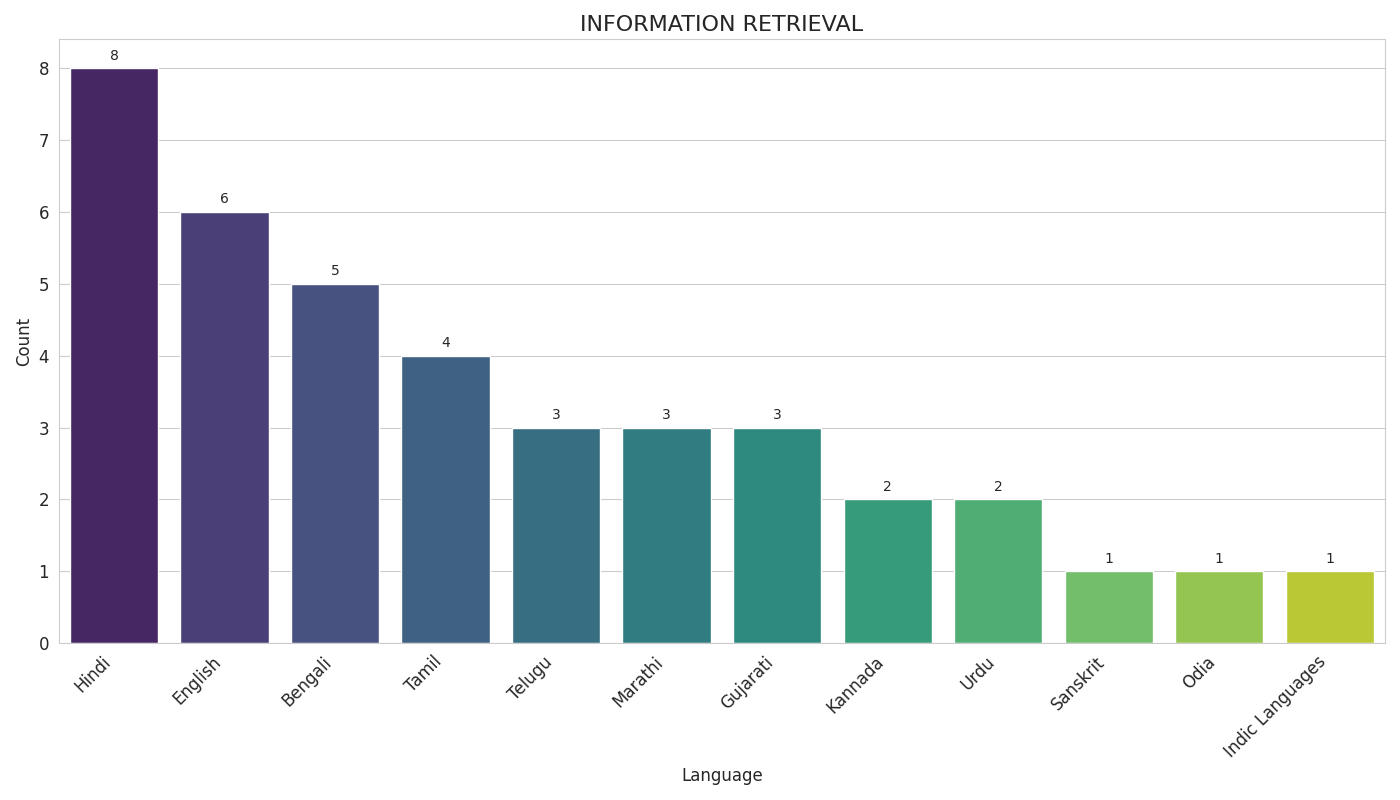}
    \caption{Language-wise distribution of datasets and studies focusing on Information Retrieval across Indian languages.}
    \label{fig:ir}
\end{figure*}

\begin{table*}[t]
\centering
\resizebox{\textwidth}{!}{
\begin{tabular}{p{3.2cm} p{3.5cm} p{2.6cm} p{4.6cm} p{4.2cm} p{4.2cm}}
\hline
\textbf{Paper} & \textbf{Focus \& Objective} & \textbf{Languages Covered} & \textbf{Methodologies \& Algorithms} & \textbf{Key Resources / Contributions} & \textbf{Key Findings} \\
\hline

\cite{jagarlamudi2007cross} &
Cross-Lingual IR (CLIR): Retrieving English documents using Indian language queries. &
Hindi, Tamil, Telugu, Bengali, Marathi, English &
SMT-based word alignment for query translation; Language Modeling for retrieval. &
Participation in CLEF 2007 Adhoc track; Monolingual and Hindi-to-English runs. &
Cross-lingual performance reached 54.4\% of monolingual; post-submission improvements raised it to 73.4\%. \\

\cite{dave2025sqcliril} &
Spoken Query CLIR benchmark for low-resource settings. &
Hindi, Gujarati, Bengali, Kannada, English &
Sparse (BM25), Dense bi-encoders, Hybrid RRF, LLM-based pointwise fusion (LPF). &
SqCLIRIL benchmark with spoken queries; human-translated TREC DL’19/20 queries. &
LPF consistently improves nDCG; generative alignment benefits speech-centric CLIR. \\

\cite{bhattacharya2018using} &
CLIR query translation using word embeddings. &
Hindi, Bengali, Marathi, Gujarati, Tamil, English &
Multilingual word embeddings; graph-based clustering using Louvain algorithm. &
Tool and visualizations for multilingual word communities. &
Multilingual clustering improves query translation precision; auxiliary languages help define dense subclusters. \\

\cite{sourabh2012query} &
Monolingual IR for Hindi addressing low recall. &
Hindi &
Query optimization strategies handling morphology and compound words. &
Analysis of linguistic challenges in Hindi IR. &
Standard IR systems suffer low recall if used without language-aware optimization. \\
\hline
\end{tabular}
}
\caption{Overview of Information Retrieval research for Indian languages, covering monolingual, cross-lingual, mixed-script, spoken-query, ontology-based, and spatial document retrieval.}
\label{tab:indic_ir_survey}
\end{table*}

\begin{figure*}[htbp]
    \centering
    \includegraphics[width=\linewidth]{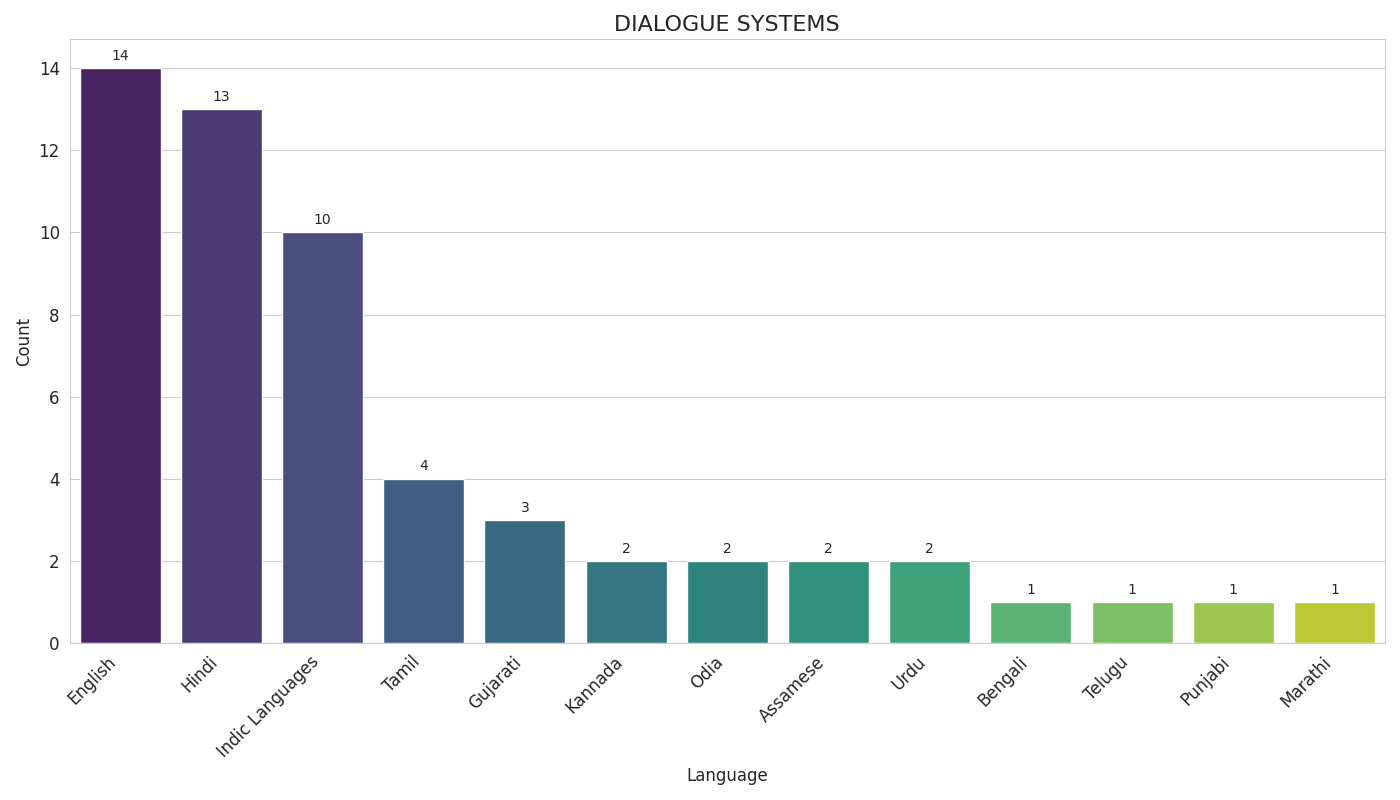}
    \caption{Language-wise distribution of datasets and studies focusing on Dialogue Systems across Indian languages.}
    \label{fig:DIALOGUE}
\end{figure*}

\begin{table*}[t]
\centering
\resizebox{\textwidth}{!}{%
% %
}
\caption{Representative dialogue system datasets, models, and applications for Indian languages.}
\label{tab:indic_dialogue_systems}
\end{table*}

\begin{figure*}[htbp]
    \centering
    \includegraphics[width=\linewidth]{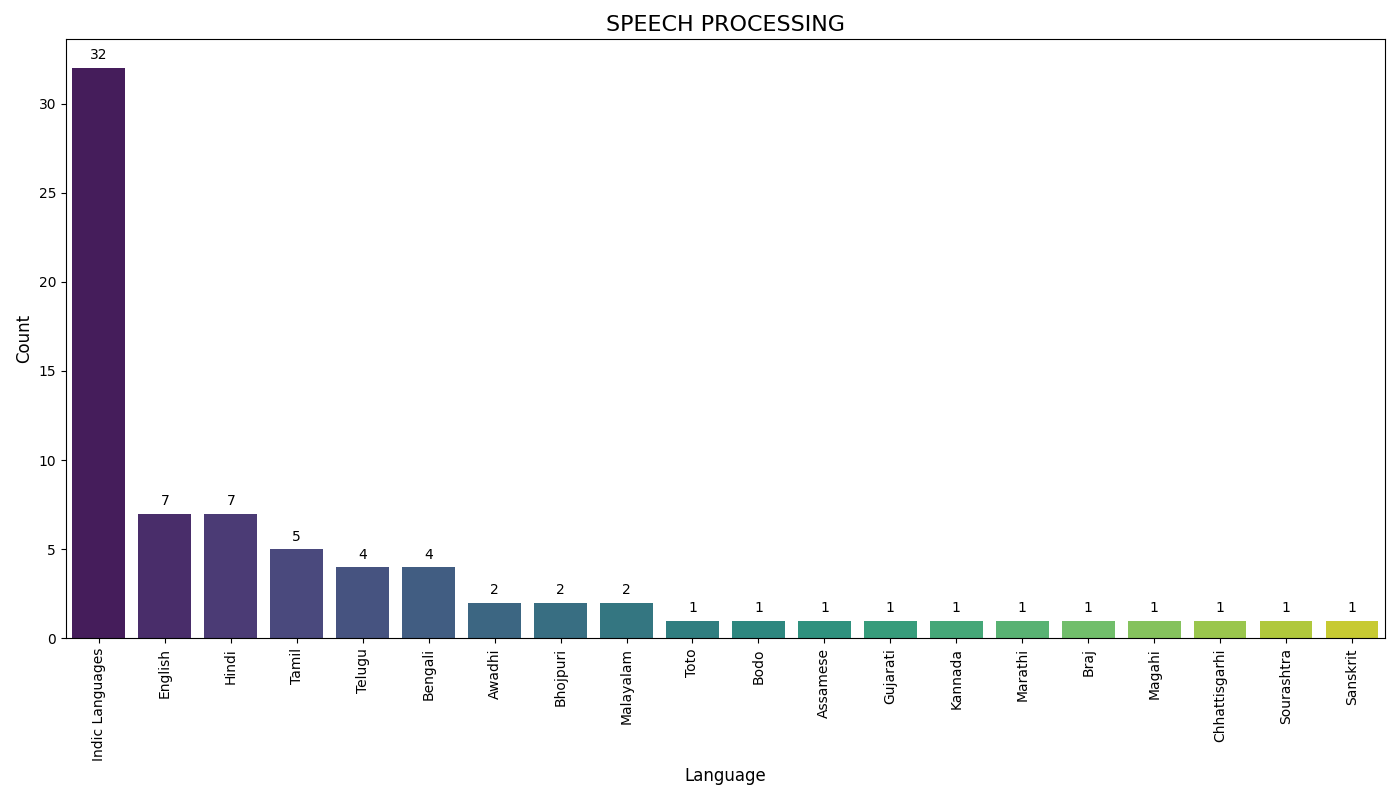}
    \caption{Language-wise distribution of datasets and studies focusing on Speech Processing Systems across Indian languages.}
    \label{fig:speech}
\end{figure*}

\begin{table*}[t]
\centering
\resizebox{0.9\textwidth}{!}{
\begin{tabular}{p{3.5cm} p{4.2cm} p{3.5cm} p{5.5cm} p{6cm}}
\hline
\textbf{Paper} & \textbf{Focus \& Objective} & \textbf{Languages Covered} & \textbf{Methodologies \& Algorithms} & \textbf{Key Resources / Contributions} \\
\hline
IndicSUPERB \cite{javed2023indicsuperb} & Benchmarking Speech Language Understanding (SLU) & Indic languages & Extension of SUPERB; self-supervised speech models (wav2vec 2.0) & IndicSUPERB benchmark; Kathbath dataset \\
IndicVoices \cite{javed2024indicvoices} & Inclusive multilingual natural speech collection & 22 languages (145 districts) & Open-source data collection blueprint; standardized protocols; QC pipelines & 7,348 hrs speech; 16,237 speakers (read, extempore, conversational) \\
IndicVoices-R \cite{sankar2024indicvoices} & Scaling TTS via enhanced ASR data & 22 Indian languages & Cross-lingual denoising and speech enhancement & 1,704 hrs high-quality speech; IV-R benchmark \\
Dementia Speech \cite{vekkot2023dementia} & Pilot dementia speech dataset & Telugu, Tamil, Hindi & Manual translation from DementiaBank; prosodic feature analysis & Indic dementia speech dataset (non-clinical) \\
IndicSpeech \cite{srivastava2020indicspeech} & TTS corpus creation & Hindi, Malayalam, Bengali & Neural TTS system training & 24-hour public TTS corpus; released models \\
BhasaAnuvaad / IndicSeamless \cite{sankar2025towards} & Large-scale speech translation & 14 Indian languages + English & Web crawling; synthetic disfluencies; ST model training & 44k hrs audio; 17M aligned segments \\
IndicSynth \cite{sharma2025indicsynth} & Synthetic speech for deepfake detection & Low-resource Indic languages & Large-scale synthetic speech generation & Multilingual synthetic ADD dataset \\
\hline
\end{tabular}
}
\caption{Overview of speech datasets, benchmarks, and modeling efforts for Indian languages.}
\label{tab:indic_speech_resources}
\end{table*}

\begin{figure*}[htbp]
    \centering
    \includegraphics[width=\linewidth]{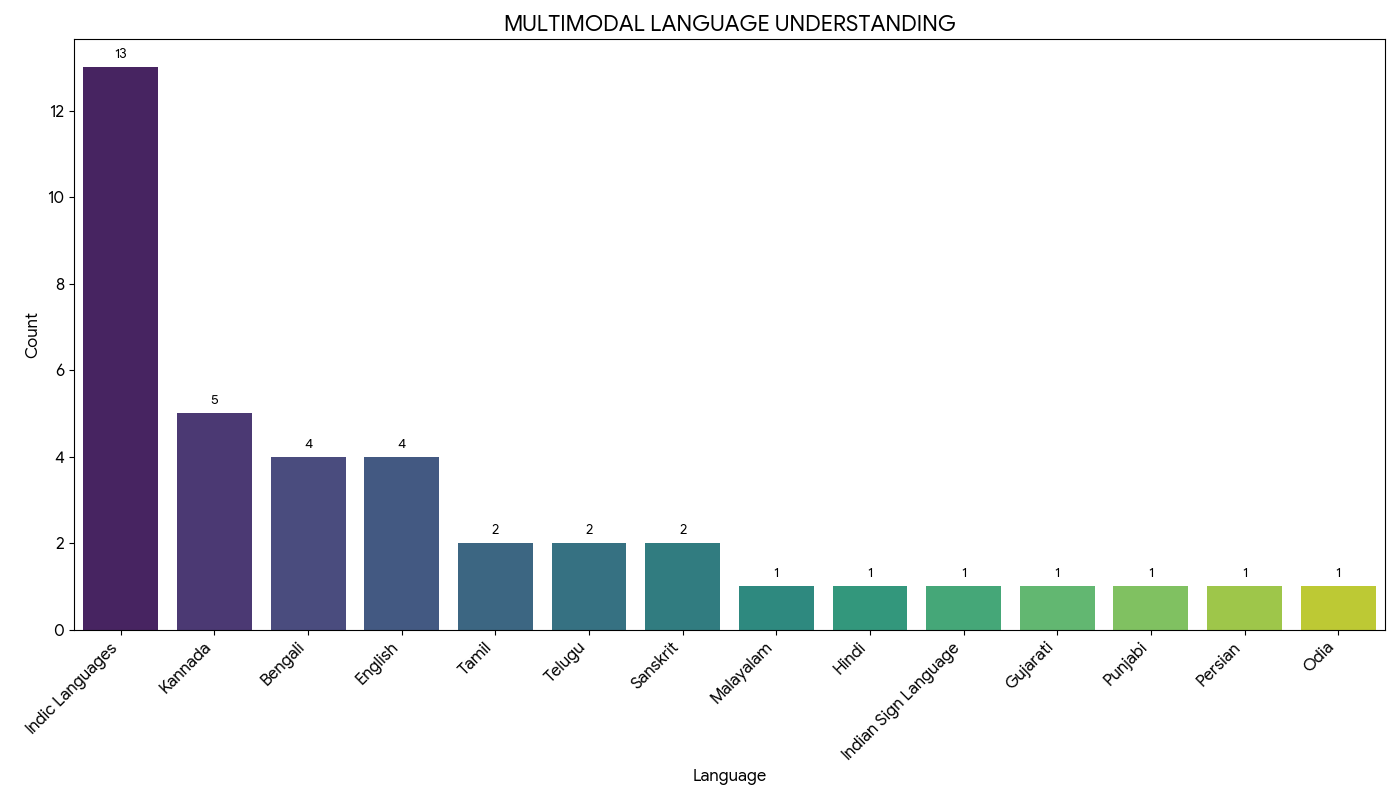}
    \caption{Language-wise distribution of datasets and studies focusing on Multimodal Language Understanding across Indian languages.}
    \label{fig:multimodal}
\end{figure*}

\begin{table*}[t]
\centering
\resizebox{\textwidth}{!}{
\begin{tabular}{p{2.6cm} p{3.2cm} p{2.8cm} p{4.2cm} p{4.2cm} p{3.8cm}}
\hline
\textbf{Paper} & \textbf{Focus \& Objective} & \textbf{Languages Covered} & \textbf{Methodologies \& Algorithms} & \textbf{Key Resources / Contributions} & \textbf{Key Findings} \\
\hline
\cite{khan2025chitrakshara} & Large-scale multilingual multimodal pretraining for multi-image reasoning. & 11 Indian languages & Data curation, filtering, and processing from Common Crawl. & Chitrakshara-IL (193M images) and Chitrakshara-Cap (44M image-text pairs). & Addresses representation gap in Indian VLMs; high diversity for Indic languages. \\

\cite{sen2022bengali} & English--Bengali multimodal MT and image captioning. & Bengali, English & Manual translation of Visual Genome captions; image-based ambiguity resolution. & 29K training segments and 1.4K challenge test set. & Images resolve ambiguities that text alone cannot. \\

\cite{chakravarthi2019multilingual} & Multimodal MT for under-resourced Dravidian languages. & Tamil, Telugu, Malayalam, Kannada & MMNMT with phonetic transcription and image features. & MMDravi dataset (30K sentences). & Phonetics and images significantly improve translation quality. \\

\cite{chauhan2021m2h2} & Multimodal humor recognition in multiparty conversations. & Hindi & Text, acoustic, and visual baselines from TV episodes. & 6,191 annotated utterances from TV sitcoms. & Multimodal context is crucial for humor detection. \\

\cite{kolavi2025nayana} & Document-centric Vision--Language Models via synthetic data. & 22 languages & Multi-task data synthesis and hierarchical layout extraction. & 3M document images with VQA and retrieval subsets. & Enables OCR, layout analysis, markdown conversion, and VQA. \\

\cite{sridhar2020include} & Indian Sign Language recognition dataset. & ISL & DNNs with augmentation, encoding, and decoding variations. & 0.27M frames, 4,287 videos, 263 signs. & Achieves 94.5\% accuracy on INCLUDE-50. \\
\hline
\end{tabular}}
\caption{Representative multimodal datasets and systems for Indian languages spanning vision--language grounding, OCR, and handwriting analysis.}
\label{tab:multimodal_indic_resources}
\end{table*}

\begin{figure*}[htbp]
    \centering
    \includegraphics[width=\linewidth]{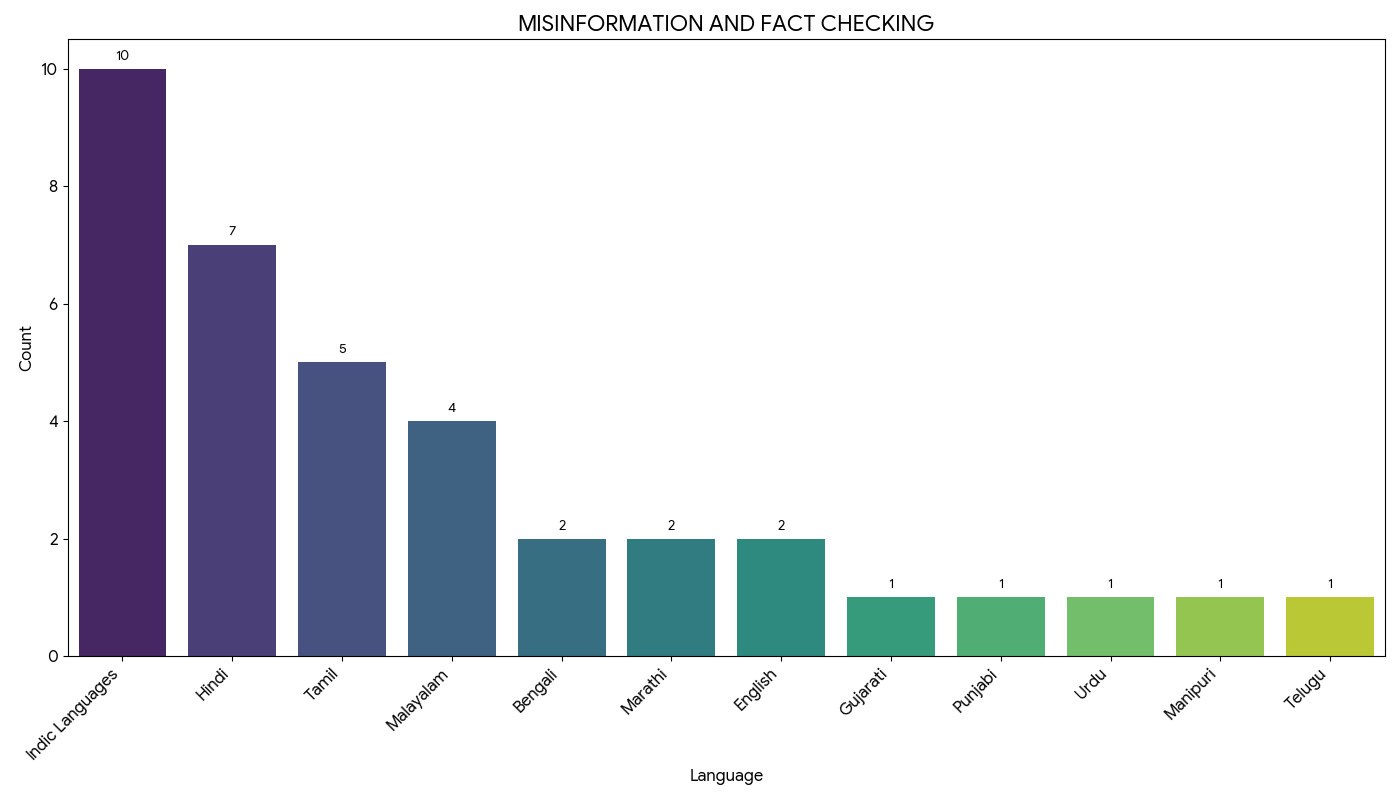}
    \caption{Language-wise distribution of datasets and studies focusing on Misinformation and Fact Checking across Indian languages.}
    \label{fig:misinfo}
\end{figure*}

\begin{table*}[t]
\centering
\resizebox{\textwidth}{!}{
\begin{tabular}{p{3.2cm} p{4.2cm} p{2.6cm} p{4.8cm} p{4.8cm}}
\hline
\textbf{Paper} & \textbf{Focus \& Objective} & \textbf{Languages Covered} & \textbf{Methodologies \& Algorithms} & \textbf{Key Resources / Contributions} \\
\hline
\cite{das2025indeepfake} & Dataset addressing deepfake detection limitations for Indian populations & Multilingual Indian & Seven multimodal manipulation techniques for deepfake creation & InDeepFake: multimodal audio--video deepfake face dataset \\
\cite{mirnalinee2022novel} & Fake news corpus creation for low-resource settings & Tamil & Web scraping; manual 5-class annotation & 2,949 fake and 2,324 genuine news samples \\
\cite{hariharan2022impact} & Evaluating transformer models for low-resource fake news detection & Tamil, Malayalam & mBERT, XLM-RoBERTa, MuRIL & Translated multilingual datasets for Tamil and Malayalam \\
\cite{kumar2025sifting} & Multimodal misinformation detection with emotional cues & Hindi & IndicBART, IndicBERT, mBERT, Vision Transformer & 6,544 article--image pairs with emotion annotations \\
\cite{sujan2023malfake} & First multimodal fake news detection study & Malayalam & RNN, VGG-16 & MALFake dataset with multiple modalities \\
\cite{francis2024tamilfacts} & Comprehensive multimodal social media analysis & Tamil & Fact-check curation; speech processing & TamilFacts: 7,934 samples incl. 884 minutes of speech \\
\cite{singhal2022factdrill} & Large-scale analysis of fake news incidents in India & 13 Indic languages & Multi-lingual, multi-media, multi-domain characterization & FactDrill: 22,435 fact-checked items (2013--2020) \\
\hline
\end{tabular}
}
\caption{Indic fake news and misinformation datasets, benchmarks, and models across multilingual and multimodal settings.}
\label{tab:indic_misinformation}
\end{table*}

\begin{figure*}[htbp]
    \centering
    \includegraphics[width=\linewidth]{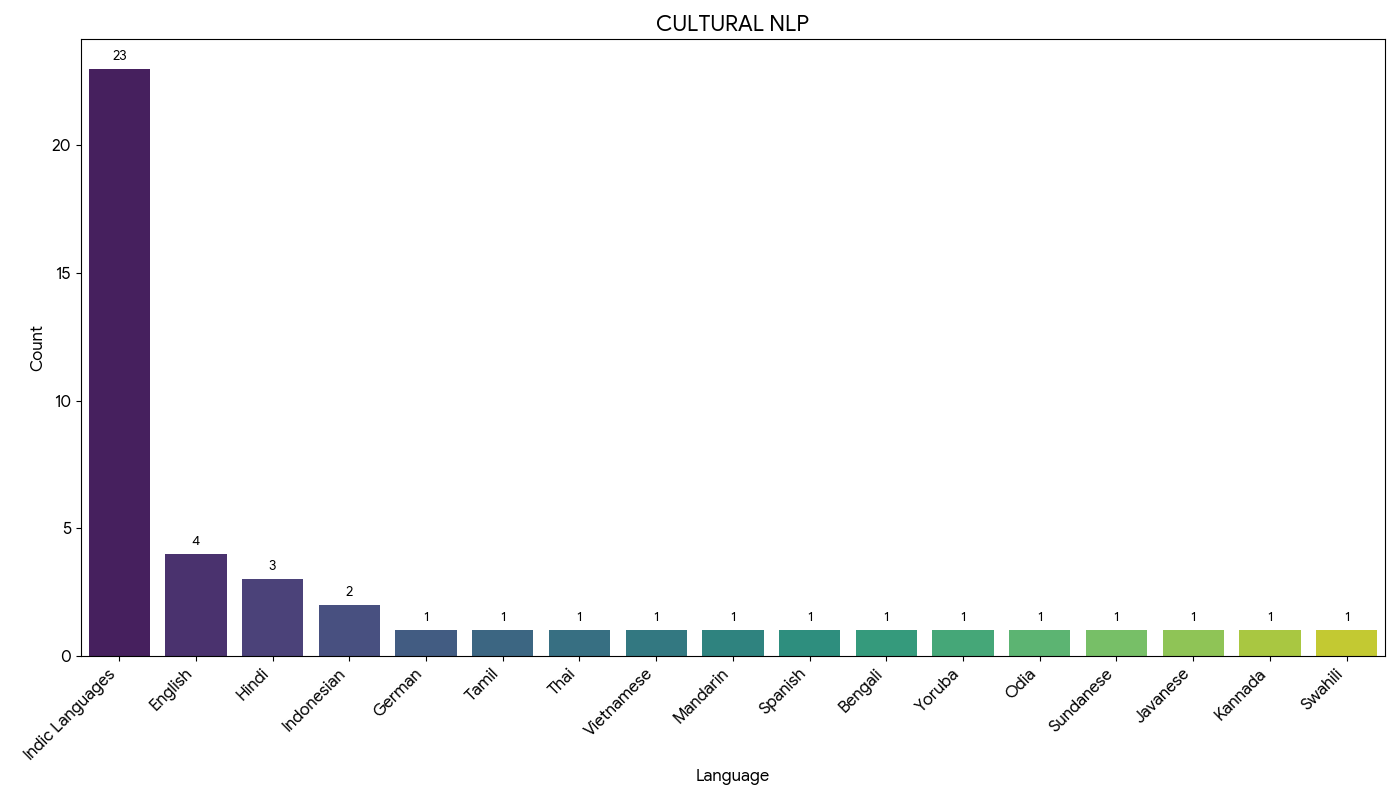}
    \caption{Language-wise distribution of datasets and studies focusing on Cultural Knowledge and Understanding across Indian languages.}
    \label{fig:cultural}
\end{figure*}

\begin{table*}[t]
\centering
\resizebox{\textwidth}{!}{
\begin{tabular}{p{2.8cm} p{4.4cm} p{3.2cm} p{4.8cm} p{4.8cm}}
\hline
\textbf{Paper} & \textbf{Focus \& Objective} & \textbf{Languages Covered} & \textbf{Methodologies \& Algorithms} & \textbf{Key Resources / Contributions} \\
\hline
\cite{watts2024pariksha} & Human--LLM agreement in multilingual, multicultural evaluation. & 10 Indic languages & Pairwise comparison, Direct Assessment; 90K human vs 30K LLM evals. & PARIKSHA leaderboards; evaluator bias analysis. \\

\cite{rai2025cross} & Cross-cultural mental health expression on social media. & English (India vs West) & Psychosocial and temporal-emotion analysis. & Linguistic differences in advice vs support seeking. \\

\cite{chhikara2025through} & LLM understanding of Indian ``Little Traditions''. & Indic languages (prompted) & Subculture case studies; prompting strategies. & Analysis of dominant vs localized narratives. \\

\cite{kirby2016d} & Global cultural, linguistic, and environmental database. & 1400+ societies & Aggregation of geo-cultural datasets. & D-PLACE database. \\

\cite{huang2023culturally} & Modeling cultural variation via NLI. & Global & Label disagreement–based NLI formulation. & CALI dataset (2.7K pairs). \\

\cite{hasan2025nativqa} & Native-user aligned cultural QA. & 7 languages & Language-agnostic native query construction. & MultiNativQA (64K QA pairs, 9 regions). \\

\cite{banerjee2025navigating} & Cultural sensitivity and harm in LLMs. & Global & Harm testing; preference fine-tuning. & Cultural harm and alignment datasets. \\
\hline
\end{tabular}}
\caption{Representative datasets, benchmarks, and models for cultural and cross-cultural NLP and multimodal understanding.}
\label{tab:cultural_nlp}
\end{table*}

\begin{figure*}[htbp]
    \centering
    \includegraphics[width=\linewidth]{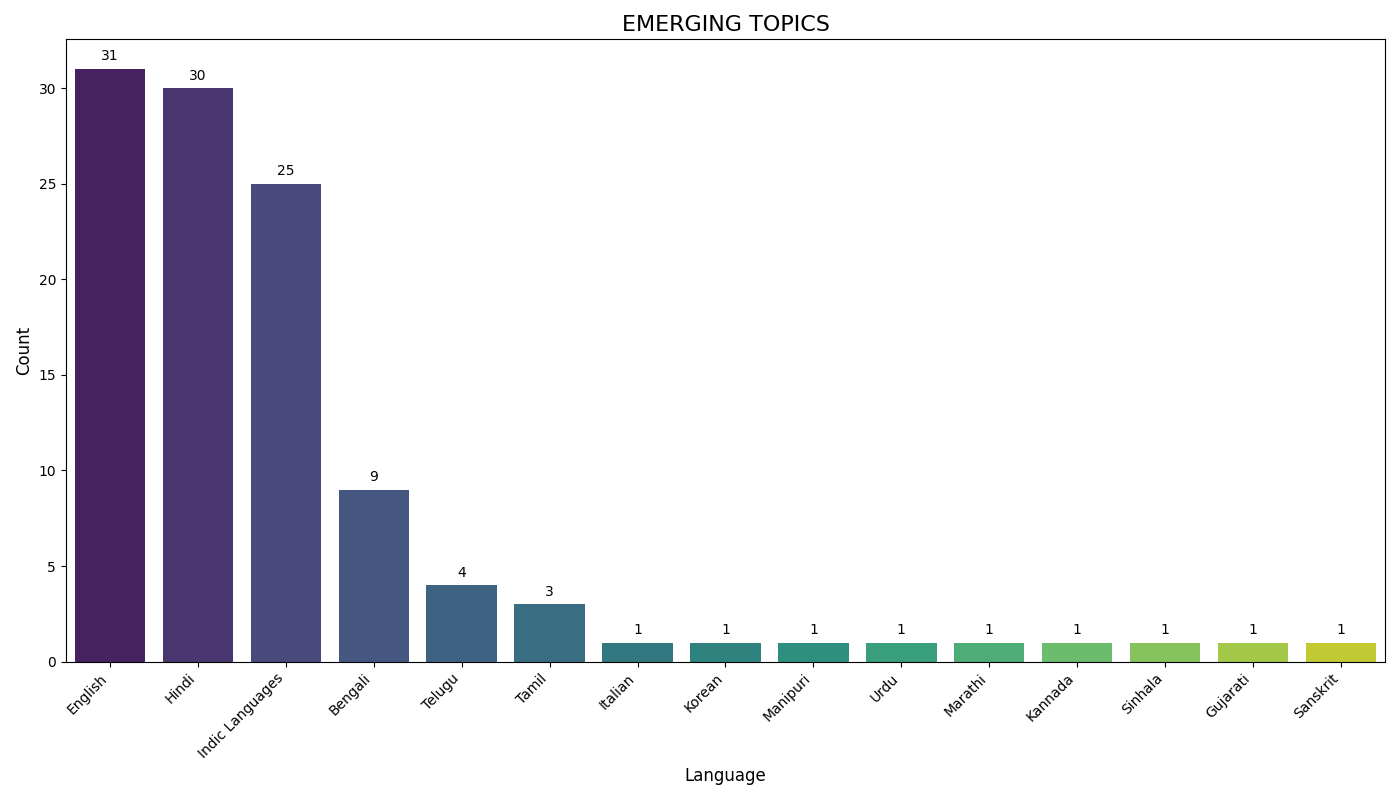}
    \caption{Language-wise distribution of datasets and studies focusing on Emerging Topics such as Bias/Fairness, Code-mixing, Style Transfer, Domain-specific Reasoning across Indian languages.}
    \label{fig:newtopics}
\end{figure*}

\begin{table*}[t]
\centering
\resizebox{\textwidth}{!}{
\begin{tabular}{p{3.2cm} p{6.4cm} p{3.6cm} p{6.0cm}}
\hline
\textbf{Paper} & \textbf{Focus \& Objective} & \textbf{Languages} & \textbf{Key Resources / Contributions} \\
\hline
\cite{sahoo2024indibias} & India-specific social bias evaluation in LLMs. & Hindi, English & IndiBias dataset (gender, caste, religion, region). \\
\cite{khandelwal2024indian} & Caste and religion stereotypes in LLMs. & English & Indian-BhED benchmark. \\
\cite{tg2025dbnlp} & Bias detection in India-centric NLP systems. & Indic, English & DBNLP bias detection framework. \\
\cite{bansal2021debiasing} & Multilingual embedding debiasing. & Hi, Bn, Te, En & State-of-the-art debiasing method. \\
\cite{tiwari2022casteism} & Casteism and gender bias in embeddings. & Hindi, Tamil & Casteism vs racism analysis. \\
\cite{malik2022socially} & Span-level social bias analysis. & Hindi & Bias encoding study. \\
\cite{das2023toward} & Cultural bias benchmarks. & Bengali & Bengali bias dataset. \\
\cite{vashishtha2023evaluating} & Gender bias in MLMs. & Multilingual & Indian-adapted gender benchmark. \\
\cite{hada2024akal} & Community-centric gender bias study. & Hindi & Field-driven dataset. \\
\cite{sahoo2023prejudice} & Bias detection in low-resource settings. & Hi, En, It, Ko & 9K annotated Hindi posts. \\
\cite{ghate2024evaluating} & Multimodal gender bias evaluation. & Hindi, English & Text–image bias framework. \\
\cite{kumar2022comma} & Aggression and bias in discourse. & Multi-Indic & ComMA dataset (59K+). \\
\cite{kamruzzaman2025banstereoset} & Social stereotypes in Bangla LLMs. & Bangla & BanStereoSet benchmark. \\
\hline
\end{tabular}}
\caption{Bias and fairness resources for Indian and multilingual NLP.}
\label{tab:bias_indic}
\end{table*}

\begin{table*}[t]
\centering
\resizebox{\textwidth}{!}{
\begin{tabular}{p{3.2cm} p{6.4cm} p{3.6cm} p{6.0cm}}
\hline
\textbf{Paper} & \textbf{Focus \& Objective} & \textbf{Languages} & \textbf{Key Resources / Contributions} \\
\hline
\cite{priyadharshini2020named} & NER for code-mixed text. & Indic CM & Script-aware NER method. \\
\cite{goswami2023offmix} & Offensive language in trilingual CM text. & Bn–En–Hi & OffMix-3L dataset. \\
\cite{nayak2022l3cube} & Pre-training on Hinglish. & Hi–En & L3Cube-HingCorpus; HingBERT. \\
\cite{hande2020kancmd} & Multi-task sentiment and offense detection. & Kn–En & KanCMD dataset. \\
\cite{smith2019language} & Code-mixed language identification. & Si–En & Sinhala–English LID dataset. \\
\cite{sheth2025comi} & Multi-task code-mixed NLP benchmark. & Hi–En & COMI-LINGUA (125K+). \\
\cite{dey2024bharatbhasanet} & Unified CM language identification. & 12 Indic & BharatBhasaNet framework. \\
\cite{bali2014borrowing} & Linguistic analysis of code-mixing. & Hi–En & Facebook CM study. \\
\cite{sandhan2022prabhupadavani} & Code-mixed speech translation. & 25 langs & 94h Vedic-domain speech corpus. \\
\hline
\end{tabular}}
\caption{Code-mixing and multilingual social media resources for Indian languages.}
\label{tab:codemix_indic}
\end{table*}

\begin{table*}[t]
\centering
\resizebox{\textwidth}{!}{
\begin{tabular}{p{3.2cm} p{6.4cm} p{3.6cm} p{6.0cm}}
\hline
\textbf{Paper} & \textbf{Focus \& Objective} & \textbf{Languages} & \textbf{Key Resources / Contributions} \\
\hline
\cite{mukherjee2024multilingual} & Multilingual sentiment style transfer. & 8 Indic & Parallel sentiment datasets. \\
\cite{krishna2022few} & Few-shot style transfer. & Multilingual & Low-resource ST model. \\
\cite{mukherjee2023text} & Toxicity removal via style transfer. & Hindi, English & Parallel detoxification dataset. \\
\cite{gunna2021transfer} & Scene text recognition. & 6 Indic & Synthetic + real image study. \\
\cite{nag2023transfer} & Relation extraction via transfer learning. & Bn, Hi, Te, En & IndoRE dataset (21K). \\
\hline
\end{tabular}}
\caption{Style transfer and controllable generation for Indian languages.}
\label{tab:style_indic}
\end{table*}

\begin{table*}[t]
\centering
\resizebox{\textwidth}{!}{
\begin{tabular}{p{3.2cm} p{6.4cm} p{3.6cm} p{6.0cm}}
\hline
\textbf{Paper} & \textbf{Focus \& Objective} & \textbf{Languages} & \textbf{Key Resources / Contributions} \\
\hline
\cite{anand2025multilingual} & Mathematical reasoning via curriculum learning. & Hindi, English & Optimized small LLMs. \\
\cite{gupta2025hats} & Analogical reasoning evaluation. & Hindi & HATS dataset. \\
\cite{bandooni2025ganitbench} & Math reasoning in VLMs. & Hindi, English & GanitBench benchmark. \\
\cite{mukherjee2025mmjee} & Multimodal scientific reasoning. & Hindi, English & mmJEE-Eval. \\
\cite{maheshwari2025parambench} & Culturally grounded graduate-level QA. & Hindi & ParamBench (17K+). \\
\cite{saxena2025riddle} & Reasoning and self-reflection. & 7 Indic & Multilingual riddles. \\
\cite{nigam2025nyayaanumana} & Legal judgment prediction. & Indic & NyayaAnumana corpus. \\
\cite{joshi2024since} & Legal reasoning benchmark. & Multi-Indic & IL-TUR leaderboard. \\
\cite{khan2025quench} & Contextual reasoning gap analysis. & English & QUENCH benchmark. \\
\cite{acharya2020towards} & Cultural commonsense reasoning. & English & Cultural Atlas dataset. \\
\hline
\end{tabular}}
\caption{Domain-specific reasoning and evaluation benchmarks for Indian NLP.}
\label{tab:reasoning_indic}
\end{table*}

\end{document}